\documentclass[letterpaper,twocolumn,10pt]{article}
\usepackage{usenix}

\usepackage{usenix,epsfig,endnotes}
\usepackage[linesnumbered,ruled]{algorithm2e}
\usepackage{balance}
\usepackage{enumitem}
\usepackage{tcolorbox}
\usepackage{diagbox}
\usepackage{xspace}
\usepackage{amsmath,amssymb,amsthm}
\usepackage{booktabs}
\usepackage{multirow}
\usepackage{makecell}
\usepackage{float}
\usepackage{authblk}

\usepackage{colortbl}  
\usepackage{xcolor, eucal}
\usepackage{array}  
\definecolor{gray0}{gray}{0.9}

\newtheorem{theorem}{Theorem}[section]
\newtheorem{definition}{Definition}[section]

\begin{document}

%don't want date printed
\date{}
\def \toolname{\textsc{Privimage}\xspace}
\def \toolnameG{\textsc{Privimage+G}\xspace}
\def \toolnameD{\textsc{Privimage+D}\xspace}

%make title bold and 14 pt font (Latex default is non-bold, 16 pt)
% \title{\Large \bf Meticulously Selecting 1\% of the Dataset for Pre-training! Generating Differentially Private Images Data with Semantics Query}
\title{\Large \bf \textsc{PrivImage}: Differentially Private Synthetic Image Generation using Diffusion Models with Semantic-Aware Pretraining}

\author[1,4,$\dag$,$\ddag$]{Kecen Li}
\author[2,$\dag$]{Chen Gong}
\author[3]{Zhixiang Li}
\author[4]{Yuzhong Zhao}
\author[1]{Xinwen Hou}
\author[2]{Tianhao Wang}

\affil[1]{Institute of Automation, Chinese Academy of Sciences}
\affil[2]{University of Virginia, $^3$University of Bristol}
\affil[4]{University of Chinese Academy of Sciences}
\affil[ ]{\{likecen2023, xinwen.hou\}@ia.ac.cn, \{chengong, tianhao\}@virginia.edu,}
\affil[ ]{ek22436@bristol.ac.uk, zhaoyuzhong20@mails.ucas.ac.cn}
% \affil[1]{likecen2023@ia.ac.cn}

\maketitle
% %
\renewcommand{\thefootnote}{\fnsymbol{footnote}}
\footnotetext[2]{Equal contribution.}
\footnotetext[3]{Work done as a remote intern at UVA.}
\renewcommand{\thefootnote}{\arabic{footnote}}

% Use the following at camera-ready time to suppress page numbers.
% Comment it out when you first submit the paper for review.
% \pagestyle{empty}

\begin{abstract}
Differential Privacy (DP) image data synthesis, which leverages the DP technique to generate synthetic data to replace the sensitive data, allowing organizations to share and utilize
synthetic images without privacy concerns. Previous methods incorporate the advanced techniques of generative models and pre-training on a public dataset to produce exceptional DP image data, but suffer from problems of unstable training and massive computational resource demands. This paper proposes a novel DP image synthesis method, termed \toolname, which meticulously selects pre-training data, promoting the efficient creation of DP datasets with high fidelity and utility. \toolname first establishes a semantic query function using a public dataset. Then, this function assists in querying the semantic distribution of the sensitive dataset, facilitating the selection of data from the public dataset with analogous semantics for pre-training. Finally, we pre-train an image generative model using the selected data and then fine-tune this model on the sensitive dataset using Differentially Private Stochastic Gradient Descent (DP-SGD). \toolname allows us to train a lightly parameterized generative model, reducing the noise in the gradient during DP-SGD training and enhancing training stability. Extensive experiments demonstrate that \toolname uses only \textbf{1\%} of the public dataset for pre-training and 7.6\% of the parameters in the generative model compared to the state-of-the-art method, whereas achieves superior synthetic performance and conserves more computational resources. On average, \toolname achieves 6.8\% lower FID and 13.2\% higher Classification Accuracy than the state-of-the-art method. The replication package and datasets can be accessed online\footnote{\url{https://dp-image-syn.github.io/privimage/}}.

%This paper highlights that \textit{"a larger pre-training dataset does not inherently lead to superior outcomes in DP image data synthesis"}, calling for attention to construct a more tailored pre-training dataset for DP dataset synthesis.

\end{abstract}

\section{Introduction}

Currently, various deep learning models increasingly rely on sensitive personal data during their training process across some crucial fields, involving healthcare~\cite{DBLP:journals/tdp/DankarE13,GONG2020107310,gong2022pattern}, finance~\cite{DBLP:journals/jpdc/HassanRC20}, and social networks~\cite{DBLP:journals/corr/abs-2010-02973}. With the rising awareness of personal privacy and data protection~\cite{AwareDP2,AwareDP3,10.1145/3564625.3564636,pang2023whitebox}, it is imperative to adopt specialized methods to enhance the security of training data, ensuring the safeguarding of users' personal information against misuse and breaches of privacy.

% In practice, our personal details, like facial information~\cite{celeba,ffhq}, can be inadvertently exposed through a sea of image data. 
% Researchers pose the question: \textit{``How can image data be used while protecting privacy?"}
Synthetic image generation with Differential Privacy (DP)~\cite{DBLP:journals/corr/abs-2104-01987} provides a potential answer to this question. 
DP image synthesis~\cite{dpsgd-gan2,dpdm-sota,dpdm,DBLP:journals/corr/abs-2112-02918,DBLP:conf/cvpr/YinMVAKM21} aims to generate synthetic images that resemble real data while ensuring the original dataset remains private. 
% In essence, the synthesized images should not disclose any specific information about an individual data point from the original dataset~\cite{}. 
% By injecting carefully calibrated noise into the synthesized images, DP ensures that sensitive information cannot be easily extracted from the generated content~\cite{DBLP:journals/csur/ZhaoC22, DBLP:journals/wicomm/LiuYZZLM21}. 
With DP image synthesis, organizations can share and utilize synthetic images, facilitating various downstream tasks without privacy concerns.

\noindent \textbf{Existing methods.} Diffusion models have demonstrated potential in DP image synthesis~\cite{ddpm,dpdm,dpldm,dpdm-sota}. 
% By modifying the standard Stochastic Gradient Descent (SGD) and adding Gaussian noise to the gradient of the model's parameters throughout training, it ensures the model does not overly adapt to specific data points or retain uncommon details that might jeopardize privacy. 
% These techniques are termed as \textit{DP-SGD diffusion}~\cite{dpdm,dpdm-sota,dpldm}. 
Dockhorn et al.~\cite{dpdm} advocated for training diffusion models using DP-SGD~\cite{dpsgd}, a widely adopted method for training models satisfying DP. 
% compact neural networks, achieving phenomenal results on a range of datasets such as {\tt Fashion-MNIST}~\cite{fmnist}, and {\tt CelebA}~\cite{celeba}. 
{Drawing inspiration from the success of pre-training and fine-tuning across many challenging tasks in computer vision~\cite{CVPretaining1,CVPretaining2,CVPretaining3,DPPublicPretraining1}, Sabra et al.~\cite{dpdm-sota} proposed to first pre-train the diffusion models on a public dataset, and then fine-tune them on the sensitive dataset.} They attained state-of-the-art (SOTA) outcomes on datasets more intricate than those used by prior methods.
% , specifically the medical image dataset {\tt Camelyon17}~\cite{camelyon1,camelyon2} and the natural image dataset {\tt CIFAR-10}~\cite{cifar10}.

% However, leveraging the entirety of a public dataset, such as {\tt ImageNet}~\cite{imagenet}, for pre-training proves inefficient. First, when the public dataset is large, we usually need networks with many more parameters for good pre-training\tw{why? why a small model doesn't work?}. Since the noise added in the DP-SGD updates is directly proportional to the number of parameters, over-parameterized models necessitate adding more noise compared to their lightly parameterized counterparts, rendering the training of diffusion DP-SGD inefficient~\cite{dpdm,nolargemodel1,nolargemodel2}. 
% % Additionally, employing large models often necessitates transferring private datasets to remote cloud services, presenting potential security concerns~\cite{DPPublicPretraining1}. 
% Secondly, not every data point within a public dataset proves beneficial for the subsequent fine-tuning through DP-SGD. Studies indicate that if there is minimal overlap between the distributions of pre-training and sensitive data, the effectiveness of pre-training diminishes~\cite{DPPublicPretraining1,DPPublicPretraining2,DPPublicPretraining3}. 

\noindent \textbf{Our Proposal.} 
% To address the aforementioned challenges, this paper presents \toolname. 
We highlight that the dataset with a semantic distribution similar to the sensitive dataset is more suitable for pre-training. 
Building on this observation, we present \toolname, an end-to-end solution to meticulously and privately select a small subset of the public dataset whose semantic distribution aligns with the sensitive one, and train a DP generative model that significantly outperforms SOTA solutions.
% This tool leverages the semantic distribution queried from the sensitive dataset to evaluate the utility of individual data points in the public dataset and select data for pre-training diffusion models, ensuring that the distribution of the selected pre-training dataset more closely aligns with that of the sensitive dataset. 
% \toolname allows us to select the most pertinent data from the public dataset for pre-training.

\toolname consists of three steps. Firstly, we derive a foundational semantic query function from the public dataset. This function could be an image caption method~\cite{ImageCaption1,ImageCaption2} or a straightforward image classifier. Secondly, \toolname uses the semantic query function to extract the semantics of each sensitive image. The frequencies of these extracted semantics then shape a semantic distribution, which can be used to select data from the public dataset for pre-training. To make this query satisfy DP, we introduce Gaussian noise to the queried semantic distribution. Finally, we pre-train image generative models on the selected dataset and fine-tune pre-trained models on the sensitive dataset with DP-SGD~\cite{dpsgd}. 

Compared to previous studies~\cite{dpdm-sota,api}, \toolname employs a more compact public dataset for pre-training, which conserves not only computational resources and time but also achieves competitive synthesis performance in terms of both fidelity and utility.

\noindent \textbf{Evaluations.} 
We conduct comprehensive experiments to verify the effectiveness of \toolname. 
% The data selected from the public dataset based on semantic distribution is effective for pre-training. 
By utilizing just \textbf{1\%} of the {\tt ImageNet}~\cite{imagenet} dataset for pre-training, we can achieve superior synthesis performance compared to existing solutions that use the full dataset for pre-training. Specifically, under three privacy budgets $\varepsilon=\{1,5,10\}$, \toolname outperforms all baselines in terms of utility and fidelity of synthetic images. On average, the FID and Classification Accuracy of the downstream classification task of synthetic images from \toolname is 30.1\% lower and 12.6\% higher than the SOTA method, i.e., PDP-Diffusion~\cite{dpdm-sota}. Besides, \toolname achieves competitive high-quality image synthesis, while just costing 50\% lower GPU memory and 48\% lower running time compared to the SOTA method.

Our experiments further explore the factors that make the pre-training dataset selected by \toolname effective. The results reveal that the success of \toolname can be attributed to two aspects: (1) \textit{Win at the starting.} \toolname selects a pre-training dataset that more closely resembles the sensitive data than the entire public data.
Before fine-tuning on the sensitive dataset, \toolname inherently produces synthetic images with a data distribution more aligned with the sensitive data than those generated by existing methods. (2) \textit{Reduce the size of the model.} Previous studies have demonstrated that as the dataset size decreases, the performance gap between large models and lightweight models narrows~\cite{scalinglaw1,scalinglaw2}. Therefore, with the reduction in the size of the pre-training dataset, \toolname enables us to well train generative models with fewer parameters, enhancing training efficiency. Additionally, we observe that the privacy budget used in selecting pre-training data is negligible, exerting minimal impact on the fine-tuning of generative models.

We also analyze the effects of hyper-parameters, specifically the selection ratio and model size, on our method. We observed that \toolname, with a selection ratio of less than 50\%, typically outperforms all baselines that use the entire public dataset. However, the performance of \toolname declines as the selection ratio rises. When using over-parameterized models, \toolname also experiences inconsistent performance. {Our study highlights that \textit{a larger pre-training dataset does not inherently lead to superior outcomes in DP image data synthesis}, emphasizing the need for constructing a more tailored pre-training dataset for DP image dataset synthesis.}

In summary, our contributions are three-fold:
\begin{itemize}[leftmargin=*]
    \item {We analyze the pre-training and fine-tuning paradigm for DP image synthesis and made the important observation that the distribution of the public data utilized should be similar to that of the sensitive data.}
    \item {We introduce \toolname, which uses the semantic distribution of the dataset requiring protection to meticulously select pre-training data. The selected dataset more closely aligns with the sensitive data compared to the entire public dataset, thereby enabling efficient synthesis of DP image datasets with high fidelity and utility.}
    \item Extensive experiments present that by utilizing only 1\%  of the public dataset for pre-training, we can significantly save computational resources and time. Specifically, the diffusion model in \toolname involves a mere 7.6\% of parameters in PDP-Diffusion, but achieves superior synthesis performance. \toolname achieves SOTA results on {\tt CIFAR-10}~\cite{cifar10} and {\tt CelebA}~\cite{celeba}.
\end{itemize}

% \noindent \textbf{Roadmap.} Section~\ref{sec:background} briefly introduces the concept of DP and diffusion models, and gives an overview of existing solutions. Section~\ref{sec:method} describes our \toolname in more details. We present the experiment setup in Section~\ref{sec:experiment_setup}. The experiment results are presented in Section~\ref{sec:rq} and Section~\ref{sec:dis}, respectively. Section~\ref{sec:related_work} discusses related works,
% after which we conclude our paper in Section~\ref{sec:conclusions}. 

\section{Background}
\label{sec:background}
This section introduces the concepts of Differential Privacy, delves into image generative models, and provides an overview of the DP image dataset synthesis techniques.

\subsection{Differential Privacy}
Differential privacy~\cite{dp} is a privacy-preserving concept quantifying sensitive data disclosure. 
% It provides a valid method to safeguard the privacy of individuals within datasets utilized for various fields, especially in machine learning. 

 \begin{definition}
 \label{def:dp}
    (Differential Privacy~\cite{dp}) A randomized algorithm $M$ satisfies ($\varepsilon, \delta$)-differential privacy, where $\varepsilon > 0$ and $\delta > 0$, if and only if, for any two adjacent datasets $D$ and $D'$, it holds that,
\begin{equation}
    \Pr[M(D) \in O] \leq e^\varepsilon \Pr[M(D') \in O] + \delta,
    \nonumber
\end{equation}
where $O$ denotes the set of all possible outputs of the algorithm $M$. The privacy budget $\varepsilon$ is a non-negative parameter that measures the privacy loss in the data. A smaller $\varepsilon$ indicates better privacy. In this study, two datasets ${D,D'}$ are deemed adjacent, denoted $D\simeq D'$, if $D=D' + x$ or $D' = D + x$, where $D' + x$ is the dataset derived by appending a data entry $x$ to dataset $D$.
\end{definition}

A popular mechanism is the Sub-sampled Gaussian Mechanism (SGM)~\cite{sgm}. Let $f:{D_s} \subseteq D \to {\mathbb{R}^d}$ be query function with sensitivity ${\Delta _f} = \max_{D\simeq D'}{\left\| {f\left( D \right) - f\left( {D'} \right)} \right\|_2}$.  SGM is parameterized with a sampling
rate $q \in \left( {0,1} \right]$ and noise standard deviation $\sigma>0$, and is defined as,
\begin{equation}
    SG{M_{f,q,\sigma }}\left( D \right) \buildrel \Delta \over = f\left( S \right) + \mathcal{N} \left( {0,{\sigma ^2}\Delta _f^2{\rm I}} \right)
    \nonumber
\end{equation}
% \tw{$M$ is a general DP mechanism, SGM is one such $M$. within SGM (and also when defining sensitivity), you shouldn't use $M$}
where $S =$ \{${x\left| x \in D \right.}$ selected independently with probability $q$\} and we define $f\left( \emptyset  \right) = 0$. We suppose that $\Delta _f^2=1$ for simplicity, which can be easily controlled by changing the $\sigma^2$. R\'{e}nyi Differential Privacy (R\'{e}nyi DP)~\cite{sgm} is usually used to track the privacy cost of SGM.
\begin{definition}
\label{def:rdp}
    (R\'{e}nyi DP~\cite{sgm}) Let ${D_\alpha }\left( {Y\left\| N \right.} \right) = \frac{1}{{\alpha  - 1}}\ln {\mathbb{E}_{x\sim N}}{\left[ {\frac{{Y\left( x \right)}}{{N\left( x \right)}}} \right]^\alpha }$ be R\'{e}nyi divergence with $\alpha>1$, a randomized mechanism $M$ is said to be ($\alpha, \gamma$)-RDP, if ${D_\alpha }\left( {M(D)\left\| M(D') \right.} \right) < \gamma$ holds for any adjacent dataset $D$, $D'$.
\end{definition}
\begin{theorem}
\label{the:rdp4sgm}
    (RDP for SGM~\cite{sgm}) Let $p_0$ and $p_1$ denote the PDF of $\mathcal{N}(0,\sigma^2)$  and $\mathcal{N}(1,\sigma^2)$ respectively. A $SG{M_{M,q,\sigma }}\left( D \right)$ satisfies ($\alpha, \gamma$)-RDP for any $\gamma$ such that,
    \begin{equation}
    \label{eq:rdp_gamma}
    \gamma  \ge {D_\alpha }\left( {\left[ {\left( {1 - q} \right){p_0} + q{p_1}\left\| {{p_0}} \right.} \right]} \right)
    \end{equation}
\end{theorem}

The above theorem shows that the privacy bound $\gamma$ can be computed using the term ${D_\alpha }\left( {\left[ {\left( {1 - q} \right){p_0} + q{p_1}\left\| {{p_0}} \right.} \right]} \right)$. 

In the domain of machine learning, the most popular approach to satisfy DP is DP-SGD\cite{dpsgd}, which modifies the standard Stochastic Gradient Descent (SGD) and adding Gaussian noise to the gradient of the model's parameters throughout training, and then updated model via,
\begin{align*}
% \vspace{-5mm}
\small
\theta \gets \theta - \eta \left( \frac{1}{|b|} \sum_{i \in b} \text{Clip}\left(\nabla {\mathcal{L}}(\theta, x_i), C\right) + \frac{C}{|b|} \mathcal{N}(0, \sigma^2 I) \right),
% \vspace{-5mm}
\end{align*}
where $\eta$ is the learning rate, $\nabla {\mathcal{L}}(\theta, x_i)$ is the gradient of the loss function ${\mathcal{L}}$ with respect to model parameters $\theta$ for the data point $x_i$ in a randomly sampled batch $b$. $\text{Clip}(\nabla {\mathcal{L}}, C)$ refers to a function that clips the gradient vector $ \nabla \mathcal{L} $ such that its $\ell_2$ norm under the constraint of $C$, and \( \mathcal{N}(0, \sigma^2 I) \) is the Gaussian noise with the variance $\sigma$. DP-SGD ensures the model does not overly adapt to specific data points or retain uncommon details that might jeopardize privacy. We provide more details in Algorithm~\ref{alg:fine-tuning} of Section~\ref{subsec:pf}. 
% \kc{We also give more details about how to choose a proper privacy budget to defend some known attacks in Appendix~\ref{ap:epsChoice}.}

\subsection{Image Generative Models}

This subsection presents two of the most effective image generative models: diffusion models~\cite{ddpm_song} and Generative Adversarial Networks (GANs)~\cite{gan}.

\noindent \textbf{Diffusion Models:} Diffusion Models~\cite{ddpm,ddpm_song,iddpm} are a class of likelihood-based generative models that consist of two processes: (1) The \emph{forward diffusion process} that progressively corrupts a clean image $x_0$ by adding Gaussian noise, which generates gradually noisier images $\{{x_1}, \ldots ,{x_T}\}$ , and $T$ is the number of noising steps. (2) The \emph{reverse process} that progressively denoise a noise to a clean image via a trainable neural network. The forward diffusion process between adjacent noisier images, i.e., $x_{t-1}$ and $x_t$, follows a multi-dimensional Gaussian distribution, which is formulated as,
\begin{equation}
\label{DiffusionProcess}
    h\left( {{x_t}\left| {{x_{t - 1}}} \right.} \right) = \mathcal{N}\left( {{x_t};\sqrt {1 - {\beta _t}} {x_{t - 1}},{\beta _t}{\rm I}} \right)
    \nonumber
\end{equation}
where ${\beta _t}$ regulates the variation of the noise distribution at each step and can be either fixed or learnable. We note ${{\bar \alpha }_t}: = \prod\nolimits_{s = 1}^t {\left( {1 - {\beta _s}} \right)} $. The likelihood between the clean image $x_0$ and noisier images in step $t$ is formulated as,
\begin{equation}
    h\left( {{x_t}\left| {{x_0}} \right.} \right) = \mathcal{N}\left( {{x_t};\sqrt {{{\bar \alpha }_t}} {x_0},\left( {1 - {{\bar \alpha }_t}} \right){\rm I}} \right)
    \nonumber
\end{equation}
Therefore, we can sample ${x_t}$ from $h\left( {{x_t}\left| {{x_0}} \right.} \right)$ directly from $x_0$ in closed form instead of adding noise $t$ times as,
\begin{equation}
    {x_t} = \sqrt {{{\bar \alpha }_t}} {x_0} + e\sqrt {1 - {{\bar \alpha }_t}} ,e\sim \mathcal{N}\left( {0,{\rm I}} \right)
    \nonumber
\end{equation}
The final objective of diffusion models is defined as by~\cite{ddpm},
\begin{equation}
\label{eq:L_DM}
{\mathcal{L}_{DM}}: = {\mathbb{E}_{t \sim U \left( {1,T} \right),{x_t} \sim h\left( {{x_t} \left| {{x_0}} \right.} \right),e \sim \mathcal{N}\left( {0,{\rm{I}}} \right)}}{\left\| {e - {e_\theta }\left( {{x_t},t} \right)} \right\|^2}
\end{equation}
where ${e_\theta }$ is a denoising network parameterized with $\theta$. Through minimizing Eq.~\eqref{eq:L_DM}, ${e_\theta }$ learns to predict the noise $e$ of any noisy images $x_t$. Thus, we can use the noise ${e_\theta }\left( {{x_t},t} \right)$ predicted by $e_\theta$ to denoise the noisy images~\cite{ddpm_song,ddpm}. After being trained well, ${e_\theta }$ can be used to gradually denoise a random Gaussian noise to a clean image.

\noindent \textbf{Generative Adversarial Nets (GAN):} GAN~\cite{gan,wgan} is a classical generative model and has achieved great performance on image synthesis task~\cite{biggan,stylegan,stylegan2}. GAN is composed of two networks, an image generator $Gen$ and an image discriminator $Dis$. The image generator $Gen$ receives a random noise vector and output an image. The image discriminator $Dis$ receives an image and outputs a score, which indicates how real the input image is. The $Gen$ is trained to generate more real images and the $Dis$ is trained to  distinguish whether its input image comes from the true dataset or is generated by the $Gen$. Mathematically, the objective function of GAN is,
% \begin{small}
\begin{equation}
\label{eq:L_GAN}
\begin{split}
    \mathop {\min }\limits_{Gen} \mathop {\max }\limits_{Dis} V\left( {Gen,Dis} \right) & = {\mathbb{E}_{x \sim \;q\left( x \right)}}\left[ {\log Dis\left( x \right)} \right] \\
     + {\mathbb{E}_{z \sim p\left( z \right)}} & \left[ {\log \left( {1 - Dis\left( {Gen\left( z \right)} \right)} \right)} \right]
\end{split}
\end{equation}
% \end{small}

\noindent Where $q(x)$ and $p(z)$ are the distribution of the real image and noise vector respectively. After solving the minimax two-player game between $Gen$ and $Dis$ as Eq.~\eqref{eq:L_GAN}, the well-trained $Gen$ generates images of high fidelity with the input of noise.

\begin{table}[!t]
\vspace{-10.0pt}
\small
    \centering
    \caption{A summary of existing methods and \toolname across various modules. `GM' and `DM' refer to the `Generative Model' and `Diffusion Models.' We record the trainable parameters of used models on {\tt CIFAR-10} or the number of used images in the pre-training dataset in parentheses. We detail the parameters of DP-CGAN when performing experiments on {\tt MNIST}. However, DP-CGAN did not run experiments on {\tt CIFAR-10}, as they failed to synthesize.}
    \label{tab:SumOfMethods}
    \begin{tabular}{l|cc}
    \toprule
    \diagbox{Method}{Module} & Pre-training & GM\\
    \hline
    DP-CGAN~\cite{dpsgd-gan8} & - & GAN (6.6M)\\
    DPDM~\cite{dpdm} & - & DM (1.8M)\\
    PDP-Diffusion~\cite{dpdm-sota} & ImageNet (1.3M) & DM (80M)\\
    \hline
    \rowcolor{gray0} \toolnameG & Selected Dataset (64K) & GAN (1.8M)\\
    \rowcolor{gray0} \toolnameD & Selected Dataset (64K) & DM (1.8M)\\
    \bottomrule
\end{tabular}
\end{table}

\subsection{DP Image Dataset Synthesis}
\label{subsec:framework}

With the rapid advancements in generative models and deep learning, a range of generative models have demonstrated remarkable image synthesis capabilities~\cite{gan,ddpm_song}. 
% Generative models, which are parameterized by neural networks, are typically trained by updating their parameters based on the gradient calculated from their objective functions and training data. 
In the DP dataset synthesis, a direct approach involves acquiring noisy gradient via DP-SGD~\cite{dpsgd} to update model parameters at each training iteration to ensure that the well-trained generative models satisfy DP~\cite{dpsgd-gan8,dpdm,dpdm-sota}.

DP-CGAN~\cite{dpsgd-gan8} is a representative method which applies DP-SGD to GANs. Since only $Dis$ accesses the sensitive data, they sanitize the gradient of $Dis$ using DP-SGD. DP-CGAN achieves good synthesis on {\tt MNIST}, but not as good on larger datasets. In recent years, diffusion models have been a promising image generative model which beats GANs~\cite{ddim}. Therefore, DPDM~\cite{dpdm} proposes to replace GANs with diffusion models in DP-CGAN. In particular, the authors use a large number of images for training at one iteration (i.e., large batch size) and use lightly parameterized diffusion models, getting rid of the ``curse of dimensionality''~\cite{dpdm,nolargemodel1,nolargemodel2}. DPDM achieves state-of-the-art (SOTA) synthesis on {\tt MNIST}~\cite{mnist} and {\tt FashionMNIST}~\cite{fmnist}.  While DP image synthesis methods have successfully protected the privacy of the above naive image datasets, many researchers are still working to design more effective DP image synthesis techniques for complex datasets, such as {\tt CIFAR-10}\cite{cifar10} and {\tt CelebA}~\cite{celeba}. 

As models and datasets grow increasingly complex, the pre-training and fine-tuning paradigm has become prevalent.
% the generative model with a public dataset is making rapid progress on solving this challenge~\cite{dpdm-sota,dpldm,api,dp-merf}. 
% \tw{maybe too much change this time; ideally, you describe the method, and you can name it based on your preference. currently DPDM and PDP-Diffusion are too similar; PDP-Diffusion should be more informative to indicate pertaining is leveraged}
PDP-Diffusion~\cite{dpdm-sota} proposes to exclusively pre-train the diffusion model on a public dataset and subsequently fine-tune it on the sensitive dataset, achieving SOTA image synthesis results on {\tt CIFAR-10}~\cite{cifar10}. Whereas, public datasets, like {\tt ImageNet}~\cite{imagenet},  are typically extensive and include over 1 million high-resolution images.

% To address this challenge, 
% a promising strategy\tw{not promising: large-batch only never solve the problem} is to use a large number of images in a single iteration, referred to as `batch size'~\cite{dpdm}. Yet, scaling up generative models to modern large-scale models, which typically generate high-quality images when the privacy protection is not required, still remains a significant challenge due to the slow-down training incurred by the noisy gradient. Subsequently, 
% researchers propose using the accessible public dataset (e.g., {\tt ImageNet}) to pre-train generative models~\cite{dpdm-sota}. \tw{since we only have 3 baselines, let's review them one by one here}

Table~\ref{tab:SumOfMethods} gives a comparison of the existing methods, including DP-CGAN~\cite{dpsgd-gan8}, DPDM~\cite{dpdm}, PDP-Diffusion~\cite{dpdm-sota}, and two versions of \toolname, namely \toolnameG and \toolnameD. Our method \toolname, instantiated with GAN and DM, distinguishes itself from existing methods with a smaller dataset for pre-training and a lower-parameterized model while achieving better performance.  The key idea is to align the semantic distribution of public and sensitive datasets, which will be described later.
% with a minimal privacy budget for data selection. This approach enables us to use a smaller dataset to pre-train image generative models with low-parameterized networks than other methods. 
% Various traditional DP image synthesis methods, which do not use pre-training or generative model techniques, work well on simpler datasets like {\tt MNIST}, while tend to fall short when applied to more complex datasets such as {\tt CIFAR 10}. We exclude them from our baselines but provide an overview in Section~\ref{sec:related_work}.

\section{Methodology}
\label{sec:method}

This paper focuses on exploring how to leverage the public dataset for more effective pre-training and proposes a novel DP image synthesis method, \toolname. We divide \toolname into three steps. Initially, we utilize the semantic labels from the public dataset to derive a semantic query function. Then, this function aids in querying the semantic distribution of the sensitive dataset, facilitating data selection for pre-training. Finally, we initiate pre-training of an image generative model using the selected data and then fine-tune this model on the sensitive dataset using DP-SGD.
% \tw{i wrote we made the observation about the importance of distribution similarities between pertaining and fine-tuning. can we say something (better with evaluation results) here? basically, you need to motivate why using semantic queries}
\begin{figure}[!t]
    \centering
    \setlength{\abovecaptionskip}{0pt}
    \includegraphics[width=1.0\linewidth]{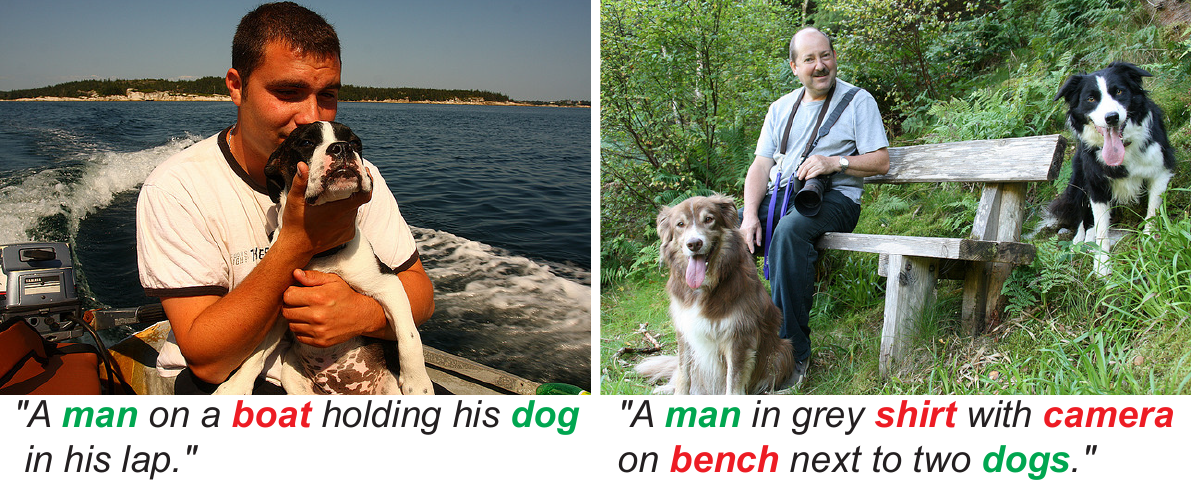}
    \caption{Two images with their captions. The same and different semantics two images own are in \textcolor[RGB]{0,165,79}{green} and \textcolor[RGB]{238,29,35}{red} respectively. Although the two images differ a lot in pixel, they have the same semantics, \textit{\textbf{\textcolor[RGB]{0,165,79}{man}}} and \textit{\textbf{\textcolor[RGB]{0,165,79}{dog}}}.}
    \label{fig:semantic_show}
\end{figure}

\subsection{Motivation}
\label{sec:motivation}

Studies indicate that if there is minimal overlap between the distributions of pre-training and target dataset (which is the sensitive dataset in our studied problem), the effectiveness of pre-training diminishes~\cite{DPPublicPretraining1,DPPublicPretraining2,DPPublicPretraining3}. Therefore, drawing inspiration from this understanding, we select the data that is ``similar'' to sensitive data from the public dataset for the pre-training of generative models to develop the DP image synthesis method. We select semantic distribution as a metric to gauge the similarity between the two datasets and provide our rationale for this choice as follows.

Semantics provide a high-level representation of images and have played a prominent role in various computer vision studies~\cite{Semantic1,Semantic2,Semantic3}. As illustrated in Figure~\ref{fig:semantic_show}, image semantics refers to the meaning and interpretation derived from an image. Unlike low-level features, such as color, texture, and edges that can be directly extracted from the image, semantics capture the “meaning” or “content” of an image, which often requires higher-level processing and understanding~\cite{Semantic4,Semantic5}. 
Most public image datasets have their semantic labels, like caption~\cite{COCOCaption}, object~\cite{COCODetection}, and category~\cite{imagenet}. 
Since our focus is on generating low-resolution images, which typically have simple content (e.g., each {\tt CIFAR-10} image contains only a single object), we propose utilizing the category labels of the public dataset to query the semantic distribution of the sensitive dataset. This information can then be used to select more relevant data for pre-training.

\subsection{Query Image Semantics}
\label{subsec:qis}

% \subsubsection{Image Semantics}
% \label{sec:image_semnatics}

% Semantics provide a high-level representation of images and have played a prominent role in various computer vision studies~\cite{Semantic1,Semantic2,Semantic3}. As illustrated in Figure~\ref{fig:semantic_show}, image semantics refers to the meaning and interpretation derived from an image. Unlike low-level features, such as color, texture, and edges that can be directly extracted from the image, semantics capture the “meaning” or “content” of an image, which often requires higher-level processing and understanding~\cite{Semantic4,Semantic5}. Most public image datasets have their semantic labels, like caption~\cite{COCOCaption}, object~\cite{COCODetection}, and category~\cite{imagenet}. We propose to leverage the semantic labels of the public dataset to query the semantic distribution of the sensitive dataset, which can be used to select more useful data for pre-training.

% Semantics provide a high-level image description, making it more comprehensible for humans. If we can extract semantic information from the sensitive dataset in a manner that conforms to DP, then we can leverage this queried semantic information to select appropriate pre-training data from the public dataset. 
We aim to derive a semantic query function that takes an image as input and returns its semantics. In this study, we use the category labels from the public dataset to represent the semantics of sensitive images. The objective while training a semantic query function $Q$ is to minimize the difference between the predicted semantic labels and the true labels. This is achieved by minimizing a loss function $\mathcal{L}_Q$, e.g., the Cross-Entropy loss~\cite{CrossEntropy}, which is defined as,

\begin{equation}
\label{eq:L_Q}
    {\mathcal{L}_Q}: = {\mathbb{E}_{\left( {x,\ {s}} \right)\sim {D_p}}}\left[ {\sum_{i = 1}^{NS} { - {s_i}\log {Q_i}\left( x \right)} } \right]
\end{equation}
where $D_p$ is the public dataset. Each pair in $D_p$ consists of an image $x$ and its semantics labels $s$. $NS$ indicates the number of possible semantics in $D_p$. $s_i = 1$ indicates that the image has the $i$-th semantic label, whereas $s_i = 0$ denotes the absence of this semantic. $Q_i(x)$ is a scalar value between 0 and 1, representing the probability that image $x$ is associated with the semantic $s_i$, as determined by the semantic query function $Q$. Through minimizing Eq.~\eqref{eq:L_Q}, we train the semantic query function $Q$ to extract semantics from images. 

After training, for each sensitive image, the semantic query function $Q$ can predict the probability that the image has the $i$-th semantic $s_i$.  Given that each image potentially can have multiple semantic labels, we select the top-$k_1$ probable semantics as predicted by $Q$ to represent the semantics of the input-sensitive image, where $k_1$ is a hyper-parameter. A higher $k_1$ suggests considering a greater number of semantics in images. 
% As illustrated in Figure~\ref{fig:semantic_show}, \kc{one} image could involve multiple semantics. We aim for the semantic query function $Q$ to characterize the input image with richer and more meaningful semantics. To achieve this, we select the top-$k$ probable semantics as predicted by $Q$ to represent the semantics of the input-sensitive image.

\begin{figure}[!t]
    \centering
    \setlength{\abovecaptionskip}{0pt}
    \includegraphics[width=1.0\linewidth]{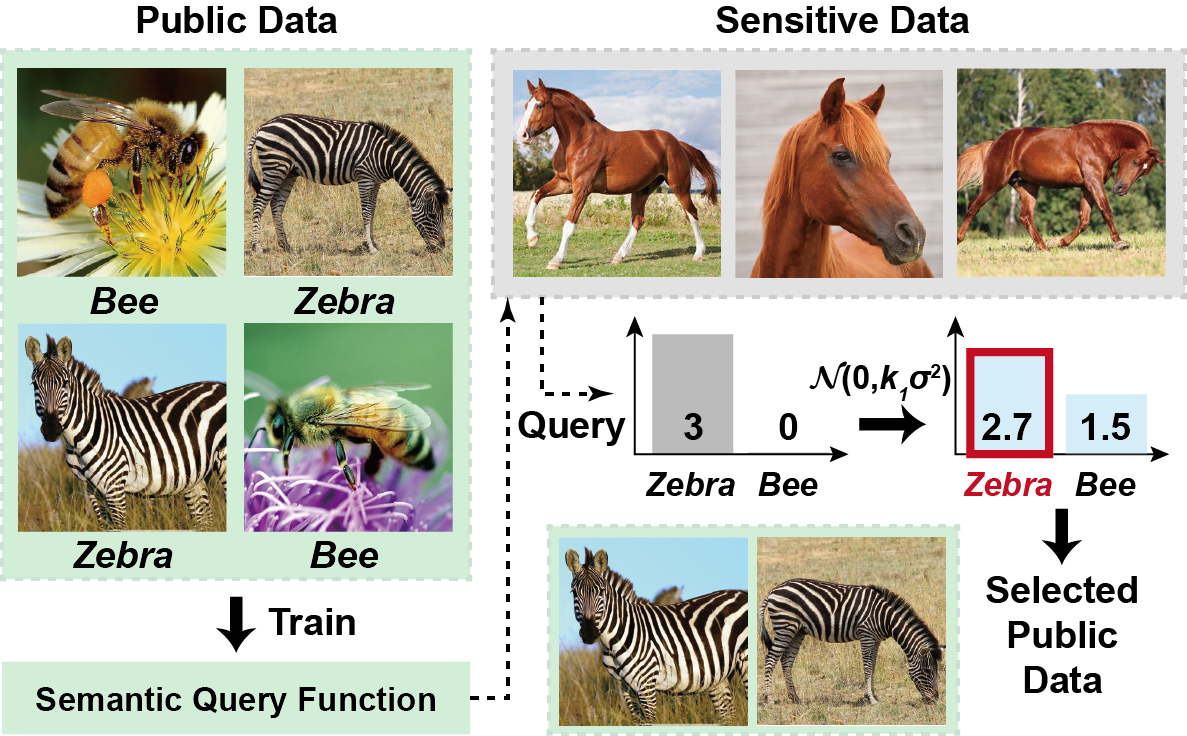}
    \caption{An example of using the semantic query function to retrieve the semantic distribution from the sensitive dataset. We first train the semantic query function using the public dataset. This function is used to obtain the semantic distribution of the sensitive dataset. To ensure privacy, we then incorporate the Gaussian noise into our query results.}
    \label{fig:sdq}
\end{figure}

\subsection{Semantic Distribution Query}
\label{subsec:sdq}
\toolname aims to select data from the public dataset for pre-training, ensuring its semantic distribution is similar to that of the sensitive dataset, without revealing any sensitive information. Towards this goal, after querying the semantics of each piece of sensitive data, this section delves into constructing a semantic distribution $SD$ for the sensitive dataset. Given that the number of semantics from the public dataset is constant, it intuitively follows that we can represent the semantic distribution as the frequency distribution of semantics derived from the sensitive dataset. Specifically, $SD$ can be conceptualized as a dictionary where the keys represent the semantics of the public dataset and the values denote the frequency of each semantic within the sensitive dataset. To safeguard the privacy of the sensitive dataset, as presented in Theorem~\ref{the:sd}, we introduce Gaussian noise to the retrieved semantic distribution, in line with the RDP (which is described in Theorem~\ref{the:rdp4sgm}), ensuring the query results adhere to differential privacy.

\begin{theorem}
    \label{the:sd}
    The semantic distribution $SD$ has global sensitivity $\Delta_{SD}=\sqrt{k_1}$, where $k_1$ is the number of semantics we query from each sensitive image. For any $\alpha>1$, incorporating noise $\mathcal{N}\left( {0,k_1{\sigma_2 ^2} {\rm I}} \right)$ into the semantic distribution $SD$ makes the query results satisfies ($\alpha$, $\alpha /\left( {2\sigma _2^2} \right)$-RDP.
    % (2) Publishing semantic distribution $SD$ with noise $\mathcal{N}\left( {0,2k\ln \frac{{1.25}}{\delta }/\varepsilon^2{\rm{I}}} \right)$ satisfies ($\varepsilon, \delta$)-DP.
\end{theorem}
We put the proof of Theorem~\ref{the:sd} in Appendix~\ref{ap:MissingProofs}. After obtaining the sensitive semantic distribution, a straightforward way is to select data from the public dataset to obtain a pre-training dataset whose semantic distribution is close to the sensitive one. However, since the sensitive semantic distribution is noisy for incorporating the Gaussian noise, it is ineffective to focus on the semantics with low frequency, which is more easily affected by the added noise. Therefore, we focus on the semantics with high frequency. Specifically, we select the top-$k_2$ semantics based on their probabilities in the semantic distribution to represent the semantic description of the sensitive dataset. 
Finally, the public data, whose semantics are included in the semantic description queried from the sensitive dataset, are selected for pre-training. Our method ensures that all semantics of the selected pre-training data fall within the high-probability regions of the sensitive semantic distribution.
% \kc{Besides, we set $k_1=k_2=k$ in following contents, which means that we use the same $k$ for semantic query in Section~\ref{subsec:qis} and semantic selection.}
We put detailed analysis in Appendix~\ref{ap:sqs}. 

We provide an example to help readers better understand the processes of semantic query and the selection of a pre-training dataset. As shown in Figure \ref{fig:sdq}, each image in the public dataset is labeled with a semantic tag, either \emph{bee} or \emph{zebra}. We first train the semantic query function using the public dataset, then determine the semantics of sensitive images using this trained function. We set $k_1=1$ and $k_2=1$, implying that each sensitive image considers only one semantic. The initial query results yield $\{zebra:3,bee:0\}$ (although the sensitive data contains horses, zebra is the closest label from $Q$).  To ensure privacy, we introduce Gaussian noise to these results, resulting in a modified output of $\{zebra:2.7,bee:1.5\}$. Given that $k_2=1$, we select the semantic \emph{zebra}, which has the highest frequency among all semantic candidates, as the representative descriptor of the sensitive dataset. Ultimately, we select only those public images labeled as \emph{zebra} for our pre-training. In a word, we select pre-training data from the public dataset guided by the perturbed semantic distribution of the sensitive dataset.

Considering applying our method for conditional generation, where the sensitive dataset has been divided into several subsets with corresponding categories, we query a semantic distribution for each subset. This query still satisfies Theorem~\ref{the:sd} because each sensitive image only provides an independent prediction. Therefore, we can label the selected image with the category of the sensitive subset it belongs to. For example, the selected samples can be zebras or, in unfortunate cases, birds from {\tt ImageNet}, but we label them as horses and pre-train our model accordingly.

% \begin{algorithm}[!t]
% 	\caption{Fine-tuning with Gradient Sanitizing}
% 	\label{alg:fine-tuning}
% 	\textbf{Input}: Sensitive dataset $D_s$; Image generative model $G$ parameterized by $\theta$; Objective of generative models: $\mathcal{L}$; Batch size $B$; Learning rate $\eta$; Scale coefficient $C$; Gaussian variance $\sigma_1^2$; Max iterations $T_m$\\
%     $T \leftarrow 0$\\
%     \While{$T<T_m$}{
%         Randomly sample the the image training batch $x_{1:B}$ from $D_s$ \\
%         \tcp{Calculate gradient}
%         $g_{1:B} \leftarrow {{\nabla _\theta } \mathcal{L} \left( {{G_\theta }\left( {{x_{1:B}}} \right)} \right)}$\\
%         \tcp{Scale gradient}
%         ${g_{1:B}}^\prime  \leftarrow \min \left\{ {1,\frac{C}{{{{\left\| {{g_{1:B}}} \right\|}_2}}}} \right\}{g_{1:B}}$\\
%         \tcp{Add Gaussian noise}
%         $\hat g \leftarrow \frac{1}{B}\sum\limits_{i = 1}^B {{g_i}^\prime }  + \frac{{\sigma_1 C}}{B}e,e\sim \mathcal{N} \left( {0,{\rm{I}}} \right)$\\
%         \tcp{Update parameter}
%         $\theta  \leftarrow \theta  - \eta \hat g$\\
%         $T \leftarrow T+1$\\
%     }
%     \textbf{Output}: The well-trained image generative model: $G^\ast$\\
% \end{algorithm}

\subsection{Pre-training and Fine-tuning}
\label{subsec:pf}

\begin{algorithm}[!t]
	\caption{Fine-tuning with Gradient Sanitizing}
	\label{alg:fine-tuning}
	\textbf{Input}: Sensitive dataset $D_s$; Image generative model $G$ parameterized by $\theta$; Objective of generative models: $\mathcal{L}$; Batch size $b$; Learning rate $\eta$; Clip coefficient $C$; Gaussian variance $\sigma_1^2$; Max iterations $T_m$\\
    $T \leftarrow 0$\\
    \While{$T<T_m$}{
        Randomly sample the the image training batch $x_{1:b}$ from $D_s$ \\
        \tcp{Calculate gradient}
        $g_{1:b} \leftarrow {{\nabla _\theta } \mathcal{L} \left( {{G_\theta }\left( {{x_{1:b}}} \right)} \right)}$\\
        \tcp{Scale gradient}
        ${g_{1:b}}^\prime  \leftarrow \min \left\{ {1,\frac{C}{{{{\left\| {{g_{1:b}}} \right\|}_2}}}} \right\}{g_{1:b}}$\\
        \tcp{Add Gaussian noise}
        $\hat g \leftarrow \frac{1}{b}\sum\limits_{i = 1}^b {{g_i}^\prime }  + \frac{{\sigma_1 C}}{b}e,e\sim \mathcal{N} \left( {0,{\rm{I}}} \right)$\\
        \tcp{Update parameter}
        $\theta  \leftarrow \theta  - \eta \hat g$\\
        $T \leftarrow T+1$\\
    }
    \textbf{Output}: The well-trained image generative model: $G^\ast$\\
\end{algorithm}

We pre-train the image generative models on the selected dataset and subsequently fine-tune these pre-trained models on the sensitive dataset. 
% We train the generative models during the pre-training stage using Stochastic Gradient Descent (SGD)~\cite{sgd}. 
In particular, in the fine-tuning phase, we use DP-SGD~\cite{dpsgd} to update the model in adherence to the DP framework, ensuring the protection of sensitive dataset's privacy. Next, we explain how to apply standard DP-SGD works in our fine-tuning in Algorithm~\ref{alg:fine-tuning}.

As described in Algorithm~\ref{alg:fine-tuning}, let $\mathcal{L}$ be the training objective of generative models (e.g., Eq.~\eqref{eq:L_DM} and~\eqref{eq:L_GAN}) and we want to minimize $\mathcal{L}$ through updating the parameters $\theta$ of our neural network $G_{\theta}$. At each training iteration, we first compute the gradient ${g = {{\nabla _\theta } \mathcal{L}\left( {{G_\theta }\left( {x} \right)} \right)}}$, where $x$ is an input image. 
% \tw{acknowledge the equation and alg1 are standard dp-sgd applied to  our setting}
Then, we clip the gradient for controlling the maximal $l_2$-norm of every gradient, like $g' = \min \left\{ {1,\frac{C}{{{{\left\| g \right\|}_2}}}} \right\} g$, where $C$ is a hyper-parameter. Lastly, we introduce Gaussian noise with variance $\sigma_1^2$ to the scaled gradient. In a word, we update the $G_\theta$ with the noisy gradient, which is defined as,
\begin{equation}
\label{eq:dpsgd}
    \hat g = \frac{1}{b}\sum\limits_{i = 1}^b {\min \left\{ {1,\frac{C}{{{{\left\| {{g_i}} \right\|}_2}}}} \right\}{g_i}}  + \frac{{\sigma_1 C}}{b}e,\;\; e\sim \mathcal{N} \left( {0,{\rm{I}}} \right),
    \nonumber
\end{equation}
\noindent {where $b$ is the batch size. Given privacy budget $\varepsilon$ and $\delta$, the maximal training iterations are constrained by the predetermined privacy budget $\varepsilon$ and $\delta$.}

\begin{algorithm}[!t]
	\caption{\toolname Workflow}
	\label{alg:privimage}
	\textbf{Input}: Public dataset $D_p$ with its semantics label $S_p$; Sensitive dataset $D_s$; Semantic query function $Q$ parameterized by $\phi$; Image generative model $G$ parameterized by $\theta$; Semantics query parameters $k_1,k_2$; Gaussian variance $\sigma_2^2$ for Semantic Distribution ($SD$); Privacy budget $\varepsilon$ and $\delta$.\\
	
    \tcp{Obtain semantic query function}
    Train $Q_\phi$ on $D_p$ and $S_p$ with Eq.~\eqref{eq:L_Q}\label{alg2:l2}\\
    \tcp{Query semantic distribution}
    Initiate the semantic distribution $SD$ $\leftarrow {{\rm O}_{NS \times 1}}$\label{alg2:l3}\\
    \For{$x \in D_s$}{\label{alg2:l4}
		Query $k_1$ semantics $\left[ {{s_1}, \cdots {s_{k_1}}} \right]$ from $Q\left( x \right)$\label{alg2:l5}\\
		\For{$s \in \left[ {{s_1}, \cdots {s_{k_1}}} \right]$}{\label{alg2:l6}
            Obtain the index j of $s$\label{alg2:l7}\\
            $SD\left[ j \right] +  = 1$\label{alg2:l8}
        }\label{alg2:l9}
	}\label{alg2:l10}
    $SD = SD + \mathcal{N} \left( {0,k_1{\sigma_2 ^2}{\rm{I}}} \right)$\label{alg2:l11}\\
    Semantic description $S^* \leftarrow$ top-$k_2$ semantics in $SD$\label{alg2:l12}\\
    \tcp{Choose data from the public dataset}
    Selected dataset $D_{ps} \leftarrow \varnothing$\label{alg2:l13}\\
    \For{$(x, s_p) \in D_p$}{\label{alg2:l14}
            \If{$s_p \in S^*$}{\label{alg2:l15}
			$D_{ps} = x \cup D_{ps}$\label{alg2:l16}
		}\label{alg2:l17}
	}\label{alg2:l18}
    \tcp{Pre-train and fine-tune $G_\theta$}
    Pre-train $G_\theta$ on $D_{ps}$\label{alg2:l19}\\
    % Obtain Gaussian variance $\sigma_1^2$ through Theorem~\ref{the:obtainSigma} \label{alg2:l20}\\
    % Fine-tune $G_\theta$ on $D_{s}$ following Algorithm~\ref{alg:fine-tuning}\label{alg2:l21}\\
    Fine-tune $G_\theta$ on $D_{s}$ following Algorithm~\ref{alg:fine-tuning}\label{alg2:l21}\\
    Generate $D_s^*$ by leveraging $G_\theta$\label{alg2:l22}\\
    \textbf{Output}: Synthetic dataset $D_s^*$\label{alg2:l23}\\
\end{algorithm}

% \tw{i feel this statement belongs to our methodology or at least next paragraph (we can say, we advocate a smaller model because xx); do we have experiments on large vs small models?}
A previous study has proved that model size scales sublinearly with data size~\cite{modelDataSize}. To achieve good pre-training on such large datasets, we tend to use models with larger parameters. In contrast, these models can often be over-parameterized when applied to the sensitive dataset. Whereas, we recommend to train models with fewer parameters. This is because more parameters indicate higher dimension of gradient. When the $l_2$-norm of gradient is constrained by clipping, the value of each gradient dimension decreases with the increase of the dimension. Consequently, the gradient becomes more susceptible to perturbations from consistent Gaussian noise, resulting in unstable training of generative models. Thus, it is preferable to train lightly parameterized generative models rather than over-parameterized ones~\cite{dpdm,nolargemodel1,nolargemodel2}. Moreover, with the reduction in the size of the pre-training dataset, \toolname enables us to well train generative models with fewer parameters during the pre-training and fine-tuning stages, thereby benefiting the efficiency of training.

Algorithm~\ref{alg:privimage} elaborates the process for \toolname, which produces a synthetic dataset of fidelity and utility similar to the sensitive dataset without privacy leakage. First, we train a semantic query function on a public dataset with Eq.~\eqref{eq:L_Q} (Line~\ref{alg2:l2}). Then, we use the trained semantic query function to query the semantic distribution of the sensitive dataset (Lines~\ref{alg2:l3}-\ref{alg2:l10}). After adding Gaussian noise to the query results (Line~\ref{alg2:l11}), we use it to select data from the public dataset for pre-training a generative model (Line~\ref{alg2:l12}-\ref{alg2:l19}). 
% Then, we obtain the Gaussian variance for gradient sanitizing through Theorem~\ref{the:obtainSigma} (Line~\ref{alg2:l20}) and 
Then, we fine-tune the generative model on the sensitive dataset with DP-SGD (Line~\ref{alg2:l21}). Finally, the fine-tuned model generates a synthetic image dataset (Line~\ref{alg2:l22}).  

The privacy analysis of \toolname is put into Appendix~\ref{ap:pa}. In particular, in \toolname, two processes consume the privacy budget: (1) querying the semantic distribution of the sensitive dataset and (2) querying the gradient during each training iteration while fine-tuning on the sensitive dataset. Both these processes can be viewed as compositions of multiple Sub-sampled Gaussian Mechanism (SGM) instances~\cite{sgm}. Therefore, we use RDP described in Theorem~\ref{the:rdp4sgm} to track the privacy budget cost of \toolname.

\begin{table}[!t]
\vspace{-7.0pt}
\small
    \centering
    \caption{Data split of three datasets used in our experiments.}
    \label{tab:DataSplit}
    \begin{tabular}{l|ccc}
    \toprule
    \diagbox{Dataset}{Subset}& Training& Validation& Test\\
    \hline
    {\tt CIFAR-10}& 45,000 & 5,000&10,000\\
    {\tt CelebA}& 162,770 & 19,867&19,962\\
    {\tt ImageNet}& 1,281,167 & 50,000& 100,000\\
    \hline
\end{tabular}
\end{table}

\section{Experiment Setup}
\label{sec:experiment_setup}

This section details the experimental settings to evaluate the proposed \toolname, including the investigated datasets, baselines, evaluation metrics, and implementation.

\subsection{Investigated Datasets}
\label{subsec:dataset}
We utilize ImageNet, a widely used dataset for pre-training in computer vision, along with CelebA and CIFAR-10 for DP image synthesis research. These sensitive datasets, featuring over 200K celebrity images and 60,000 natural images across 10 classes respectively, offer a greater synthesis challenge compared to the commonly used MNIST and FashionMNIST. All datasets are divided into a training set, a validation set, and a test set. Please refer to Table~\ref{tab:DataSplit} for details. All images in {\tt CelebA} undergo center-cropping and are subsequently resized to dimensions of 64×64 and 32×32, which are named {\tt CelebA32} and {\tt CelebA64} respectively. Figure~\ref{fig:dataset_samples} presents examples of {\tt ImageNet}, {\tt CIFAR-10}, and {\tt CelebA}. We present more details in Appendix~\ref{ap:dataset}.

\subsection{Baselines}
\label{subsec:baselines}
This paper selects DPDM~\cite{dpdm}, PDP-Diffusion~\cite{dpdm-sota}, DP-LDM~\cite{dpldm}, DPSDA~\cite{api}, DPGAN~\cite{dpsgd-gan3,dpsgd-gan8,dpsgd-gan9} and DPGAN with pre-training (DPGAN-p) as our baselines, which follow the same framework of DP image synthesis proposed in Section~\ref{subsec:framework}. All these baselines are trained with state-of-the-art image generative models, GANs and diffusion models, and incorporate DP-SGD~\cite{dpsgd} to ensure the protection of sensitive datasets. Additionally, PDP-Diffusion, DP-LDM, DPSDA and DPGAN-p leverage the public dataset for pre-training.

% \noindent \textbf{GS-WGAN:} The GS-WGAN model, as presented in the work of~\cite{dpsgd-gan2,dpsgd}. To regulate the sensitivity of the model during this process, the GS-WGAN leverages the properties of the Wasserstein distance~\cite{wgan}. Adopting the Wasserstein distance provides a more stable gradient flow, ensuring that the sanitized generator produces high-quality outputs while adhering to differential privacy constraints.

\begin{itemize}[leftmargin=*]

\item  \textbf{DPDM:} DPDM~\cite{dpdm} trains diffusion models with lightweight parameters and a substantial number of images (e.g., `batch sizes') at each training iteration. To protect the sensitive dataset, Gaussian noise is introduced to the gradient of the diffusion model, aligning with the DP-SGD methodology~\cite{dpsgd}.

\item \textbf{PDP-Diffusion:} PDP-Diffusion~\cite{dpdm-sota} adopts a training approach that capitalizes on larger batch sizes to enhance the stability and convergence speed of the model training. By leveraging the public dataset, the model benefits from a broader knowledge base, potentially improving its subsequent fine-tuning on the sensitive dataset.

\item \textbf{DP-LDM:} Given a pre-trained diffusion model, DP-LDM~\cite{dpldm} proposes to fine-tune only its label embedding module and attention module. This approach reduces the number of trainable parameters and requires less noise to achieve the same privacy budget.

\item \textbf{DPSDA:} DPSDA~\cite{api} proposes a Private Evolution algorithm that progressively guides the pre-trained models to generate a synthetic image dataset similar to the sensitive one without the need for fine-tuning.

\item \textbf{DPGAN:} DPGAN~\cite{dpsgd-gan3,dpsgd-gan8,dpsgd-gan9} represents a typical method that trains GAN with DP-SGD for sensitive data synthesis. We refer to the implementation of~\cite{dpsgd-gan8}, which is the sole work offering executable code along with privacy analysis. 
% We substituted their network architectures with one of the most efficient GANs -- BigGAN~\cite{biggan}.

\item \textbf{DPGAN-p:} To validate the effectiveness of dataset as selected by \toolname, we introduce an additional baseline, DPGAN-p. Building upon DPGAN, DPGAN-p first pre-trains the GAN on the public dataset, and subsequently fine-tunes it on the sensitive dataset adopting the pre-training method proposed by~\cite{dpdm-sota}.

\end{itemize}

Recently, PDP-Diffusion~\cite{dpdm-sota} with 80M parameters has achieved SOTA results on {\tt CIFAR-10}~\cite{cifar10} under $\varepsilon=10$.
For the dataset {\tt CIFAR-10}, we set $\delta$ to $10^{-5}$, and for {\tt CelebA}, we set it to $10^{-6}$. We adopt three privacy budgets, $\varepsilon=\{10,5,1\}$, for both datasets. These specific privacy budgets have been commonly used in prior DP image synthesis research~\cite{dpdm,dpdm-sota}.
\toolname prioritizes efficient image synthesis using low-parameterized neural networks even under smaller privacy budgets 5 and 1, saving substantial computation resources and achieving superior results.

\subsection{Evaluation Metrics}
\label{subsec:metrics}
We evaluate the fidelity and utility of the synthetic dataset using two widely accepted metrics: Fr\'{e}chet Inception Distance (FID) and Classification Accuracy, as commonly employed in prior research~\cite{dpdm,dpdm-sota,dpsgd-gan8}.

\noindent \textbf{Fr\'{e}chet Inception Distance (FID):} FID is a metric widely used to assess the fidelity of images generated by Generative models~\cite{ddpm,ddim,ddpm_song,biggan}. A lower FID suggests that the generated images are higher quality and more akin to the real ones. We generate 5,000 synthetic images to calculate FID.

\noindent \textbf{Classification Accuracy (CA):} We assess the utility of synthetic images with specific attributes or categories in the accuracy of downstream classification. Specifically, we select three classification models: Logistic Regression (LR), Multi-Layer Perceptron (MLP), and Convolutional Neural Network (CNN). These models are trained on synthetic datasets using different methods and their classification accuracy is tested on the sensitive test dataset. We generate 50,000 synthetic images to train classifiers.

\noindent \textbf{Semantic Distribution Similarity (SDS):} This paper proposes a metric to access the semantic distribution similarity between two image datasets. Let $D_1$ and $D_2$ be the two image datasets with semantics of $\left\{ {s_1^j} \right\}_{j = 1}^{N{S_1}}$ and $\left\{ {s_2^j} \right\}_{j = 1}^{N{S_2}}$ respectively, where $NS_1$ and $NS_2$ represent the number of semantics in their respective images. The semantic distribution similarity of $D_1$ and $D_2$ is calculated as, 

\begin{equation}
\label{eq:sds}
    SDS\left( {{D_1},{D_2}} \right) = \sum\limits_{i = 1}^{N{S_1}} {\sum\limits_{k = 1}^{N{S_2}} {w_1^iw_2^k\frac{{V(s_1^i) \cdot V(s_2^k)}}{{\left\| {V(s_1^i)} \right\|\left\| {V(s_2^k)} \right\|}}} }
\end{equation}
where $V$ indicates a function that converts a semantic or a word into its embedding vector~\cite{glove}. Besides, ${\frac{{V(s_1^i) \cdot V(s_2^k)}}{{\left\| {V(s_1^i)} \right\|\left\| {V(s_2^k)} \right\|}}}$ calculates the cosine similarity of the embedding vector of $s_1^i$ and $s_2^k$. The ${w_1^i}$ represents the frequency of semantic $s_1^i$ in $D_1$, and we consider semantics with higher frequency can better represent the dataset. The higher value of $SDS(D_1,D_2)$ suggests that $D_1$ and $D_2$ have closer semantic distribution.

\subsection{Implementation}
\label{sub:Imp}

All image generative methods are realized with Python 3.8 on a server with 4 NVIDIA GeForce A100 and 512GB memory. 

We replicate DPDM~\cite{dpdm} using open-source implementation repositories, and building upon that, we also reproduce PDP-Diffusion~\cite{dpdm-sota}. In particular, we use a larger batch size and pre-train the diffusion models on {\tt ImageNet}~\cite{imagenet}. For DP-LDM~\cite{dpldm}, we follow their proposed approach and fine-tune only the label embedding and attention modules in the diffusion models, keeping the other parameters frozen. We replicate DPSDA~\cite{api} using their open-source code and replace their RANDOM-API with the pre-trained diffusion model used in PDP-Diffusion for a fair comparison. To ensure fairness, we discuss the comparison with their 270M diffusion models in Section~\ref{subsec:ComputationalResource}. For DPGAN~\cite{dpsgd-gan8}, we refer to the practice adopted for GAN training with gradient sanitizing released in opacus\footnote{https://github.com/pytorch/opacus/tree/main/examples}. Our implementation of DPGAN-p is based on DPGAN, incorporating similar training techniques as suggested in PDP-Diffusion~\cite{dpdm-sota}. 

For \toolname, we choose diffusion models and GANs as the image generative models within our \toolname framework. We refer to these two variations as \toolnameD and \toolnameG, respectively. For the fair comparison, all Diffusion-based and GAN-based methods use lightly parameterized models with the same parameters. Although PDP-Diffusion proposes to use large-scale diffusion models (e.g. with 80M parameters), we believe this way is ineffective for both the synthesis performance and computational resource saving, which are discussed in Section~\ref{subsec:ComputationalResource} in more details. We recommend readers refer to Appendix~\ref{ap:ImpleDetails} for more implementation details.

% Additionally, we establish three other baselines: Non-private G, Non-private D, and NonPriv. Non-private G and Non-private D represent the optimal versions of \toolnameG and \toolnameD, respectively, \kc{ and gradient of models.  For NonPriv, the synthetic images are the original sensitive images.}

% \begin{table}[!t]
% \vspace{-5.0pt}
% \setlength{\tabcolsep}{3pt}
% \small
%     \centering
%     \caption{FID and CA of \toolname and four baselines on {\tt CIFAR-10}~\cite{cifar10}, {\tt CelebA32} and {\tt CelebA64}~\cite{celeba} with $\varepsilon=10$. Best performance in each column is highlighted. }
%     \label{tab:rq1}
%     \begin{tabular}{l|cccc|c|c}
%     \toprule
%     \multirow{3}{*}{Method} & \multicolumn{4}{c|}{{\tt CIFAR-10}} & {\tt CelebA32} & {\tt CelebA64 }\\
%     \Xcline{2-7}{0.5pt}
%     & \multicolumn{3}{c}{CA (\%)} & \multirow{2}{*}{FID} & \multirow{2}{*}{FID} & \multirow{2}{*}{FID}\\
%      \Xcline{2-4}{0.5pt}
%      & LR & MLP & CNN & &\\
%     \hline
%     DPGAN
% & 9.17 & 8.4  & 10.5  & 258 & 202 & 121 \\
%     DPGAN-p 
% & 13.6 & 14.9  & 24.1 & 49.7& 29.7 & 51.1 \\
%     DPDM 
% & 20.7 & 24.6  & 21.3 & 304 & 113 & 115 \\
%     PDP-Diffusion  
% & 18.7 & 21.4 & 30.4 & 66.8& 22.6 & 51.6\\
%     \hline
%     \rowcolor{gray0} \toolnameG 
% & 19.9 & 24.5 & 44.3  & 28.1& \textbf{18.9} & \textbf{38.2} \\
%     \rowcolor{gray0} \toolnameD & \textbf{32.6} & \textbf{36.5}  & \textbf{68.8} & \textbf{27.6} & 19.1 & 49.3 \\
%     \bottomrule
% \end{tabular}
% \vspace{-8.0pt}
% \end{table}

\begin{table*}[!t]
\vspace{-5.0pt}
\setlength{\tabcolsep}{2.6pt}
\small
    \centering
    \caption{FID and CA of \toolname and four baselines on {\tt CIFAR-10}~\cite{cifar10}, {\tt CelebA32} and {\tt CelebA64}~\cite{celeba} with $\varepsilon=10,5,1$. For space limitation, {\tt CeA32} and {\tt CeA64} refer to {\tt CelebA32} and {\tt CelebA64} respectively. The best performance in each column is highlighted using the bold font.}
    \label{tab:rq1}
    \begin{tabular}{l|cccc|c|c|cccc|c|c|cccc|c|c}
    \toprule
    \multirow{4}{*}{Method} & \multicolumn{6}{c|}{$\varepsilon=10$} & \multicolumn{6}{c|}{$\varepsilon=5$} & \multicolumn{6}{c}{$\varepsilon=1$}\\
    \Xcline{2-19}{0.5pt}
    & \multicolumn{4}{c|}{{\tt CIFAR-10}} & {\tt CeA32} & {\tt CeA64} & \multicolumn{4}{c|}{{\tt CIFAR-10}} & {\tt CeA32} & {\tt CeA64} & \multicolumn{4}{c|}{{\tt CIFAR-10}} & {\tt CeA32} & {\tt CeA64}\\
    \Xcline{2-19}{0.5pt}
    & \multicolumn{3}{c}{CA(\%)} & \multirow{2}{*}{FID} & \multirow{2}{*}{FID} & \multirow{2}{*}{FID} & \multicolumn{3}{c}{CA(\%)} & \multirow{2}{*}{FID} & \multirow{2}{*}{FID} & \multirow{2}{*}{FID} & \multicolumn{3}{c}{CA(\%)} & \multirow{2}{*}{FID} & \multirow{2}{*}{FID} & \multirow{2}{*}{FID}\\
     \Xcline{2-4}{0.5pt} \Xcline{8-10}{0.5pt} \Xcline{14-16}{0.5pt}
     & LR & MLP & CNN & & & & LR & MLP & CNN & & & & LR & MLP & CNN & &\\
    \hline
    DPGAN & 9.17 & 8.4  & 10.5  & 258 & 202 & 121 & 14.2 & 14.6 & 13.0 & 210 & 227 & 190 & 16.2 & 17.4 & 14.8 & 225 & 232 & 162\\
    DPGAN-p & 13.6 & 14.9  & 24.1 & 49.7& 29.7 & 51.1 & 13.9 & 14.3 & 19.2 & 48.5 & 23.9 & 52.2 & 11.3 & 13.8 & 12.8 & 70.1 & 37.9 & 54.5\\
    DPDM & 20.7 & 24.6  & 21.3 & 304 & 113 & 115 & 21.1 & 24.7 & 22.0 & 311 & 122 & 127 & 19.6 & 22.3 & 14.7 & 340 & 223 & 243\\
    DP-LDM & 15.2 & 14.1 & 26.0 & 48.6 & 21.9 & 58.0 & 15.3 & 14.6 & 24.8 & 48.9 & 22.2 & 63.9 & 12.8 & 11.8 & 18.8 & 50.1 & 45.5 & 131.9\\
    PDP-Diffusion & 18.7 & 21.4 & 30.4 & 66.8& 22.6 & 51.6 & 19.3 & 22.2 & 28.7 & 70.0 & 23.6 & 55.9 & 17.7 & 19.4 & 22.9 & 87.5 & 33.7 & 77.7\\
    DPSDA & 24.1 & 25.0 & 47.9 & 29.9 & 23.8 & 49.0 & 23.5 & 24.4 & 46.1 & 30.1 & 33.8 & 49.4 & 24.2 & 23.6 & 47.1 & 31.2 & 37.9 & 54.9\\
    \hline
    \rowcolor{gray0} \toolnameG & 19.9 & 24.5 & 44.3  & 28.1& \textbf{18.9} & \textbf{38.2} & 19.6 & 24.6 & 39.2 & 29.9 & \textbf{19.8} & \textbf{45.2} & 15.8 & 18.0 & 25.5 & 47.5 & 31.8 & \textbf{45.1}\\
    \rowcolor{gray0} \toolnameD & \textbf{32.6} & \textbf{36.5}  & \textbf{68.8} & \textbf{27.6} & 19.1 & 49.3 & \textbf{32.4} & \textbf{35.9} & \textbf{69.4} & \textbf{27.6} & 20.1 & 52.9 & \textbf{30.2} & \textbf{33.2} & \textbf{66.2} & \textbf{29.8} & \textbf{26.0} & 71.4\\
    \bottomrule
\end{tabular}
\end{table*}

\begin{figure*}[!t]
    \centering
    \includegraphics[width=0.95\linewidth]{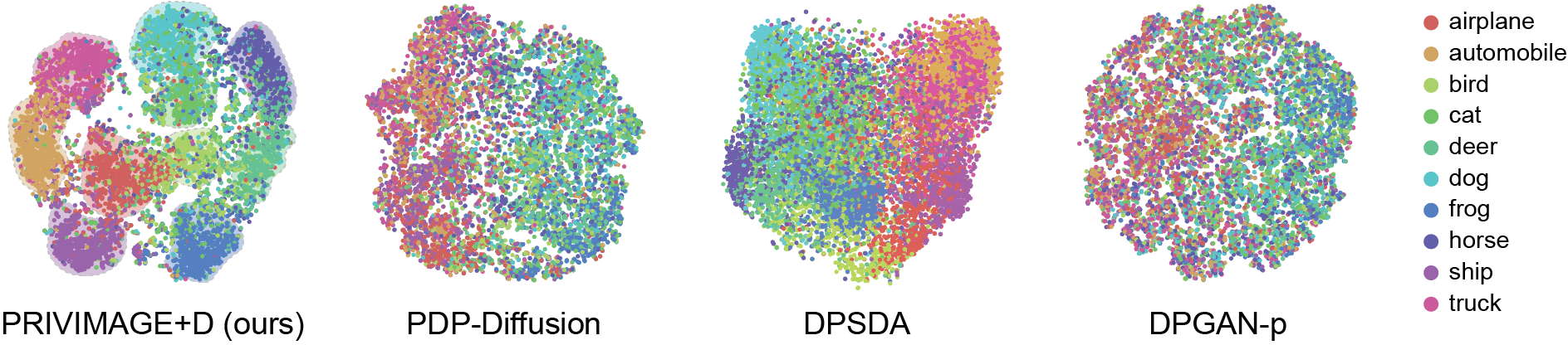}
    \caption{The t-SNE visualizations depict the embedding distribution of the synthetic images using four DP dataset synthetic methods within the {\tt CIFAR-10} dataset with $\varepsilon=10$. We obtain embeddings by using CNN classifiers trained on synthetic images.}
    \label{fig:tsne}
        \vspace{-4.0pt}
\end{figure*}

% \begin{figure*}[t]
%     \centering
%     \setlength{\abovecaptionskip}{0pt}
%     \includegraphics[width=1\linewidth]{Figures/RQ1/cifar10_sample_comp.png}
%     \caption{Examples of Synthetic {\tt CIFAR-10}~\cite{cifar10} images with $\varepsilon=10$. These generative models are trained using our tools \toolnameD, as well as PDP-Diffusion~\cite{dpdm-sota} and DPGAN-p. Each row corresponds to a category from the {\tt CIFAR-10} dataset. }
%     \label{fig:cifar10_comp_show}
%     \vspace{-4.0pt}
% \end{figure*}

\section{Evaluation Results}
\label{sec:rq}
% This section addresses four research questions (RQs) to evaluate the effectiveness of \toolname. Specifically, we investigate (1) whether \toolname generates higher-quality images than the baselines, (2) reasons why the queried semantic distribution improves fine-tuning, (3) performance of \toolname under a constrained privacy budget\tw{can we merge (3) with (1)?}, and (4) impact of hyper-parameters on \toolname.
This section addresses three research questions (RQs) to evaluate the effectiveness of \toolname. Specifically, we investigate (1) whether \toolname generates higher-quality images than the baselines, (2) reasons why the queried semantic distribution improves fine-tuning, and (3) impact of hyper-parameters on \toolname.

\subsection*{RQ1. How effective is \toolname for synthesizing useful images?}

\noindent \textbf{Experiment Design.} We explore whether \toolname can generate synthetic images with higher fidelity and utility than baselines. We compare our \toolnameG and \toolnameD (which is described in Section~\ref{sub:Imp}) with six baselines introduced in Section~\ref{subsec:baselines} on {\tt CIFAR-10}~\cite{cifar10} and {\tt CelebA}~\cite{celeba} under the privacy budget $\varepsilon=\{10,5,1\}$.

\noindent \textbf{Result Analysis.} Table~\ref{tab:rq1} shows that \toolname outperforms all baselines in terms of the FID and CA of downstream classification tasks using synthetic images on {\tt CIFAR-10}, {\tt CelebA32} and {\tt CelebA64} across three distinct privacy budgets. 
% \kc{Conditional {\tt CelebA} generation results are put into Appendix~\ref{ap:conditionalCelebA}.}

When $\varepsilon=10$, for {\tt CIFAR-10}, \toolnameG and \toolnameD outperform all the baselines in terms of FID with specifically 7.7\% lower than the SOTA method DPSDA~\cite{api}, suggesting that the fidelity of images generated by \toolname more closely aligns with the sensitive dataset. For the utility of synthetic images, \toolnameD achieves superior downstream classification accuracy on all three classification models with especially \textbf{20.9\%} higher than the SOTA method~\cite{api} on the CNN classifier. \toolnameG outperforms all GAN-based baselines and DPDM~\cite{dpdm}, achieving competitive results with the SOTA method. \toolnameG outperforms all the baselines in terms of FID with specifically 13.7\% and 22.0\% lower than the SOTA method on {\tt CelebA32} and {\tt CelebA64}, which implies that our \toolname is also effective when there is a substantial difference between the public and sensitive dataset.
Although \toolnameD does not achieve a lower FID than DPSDA on {\tt CelebA64}, we consider a practical advantage of \toolnameD to be its ability to directly sanitize the diffusion model. Once the model is well-trained, \toolname can respond to any number of synthesis queries with fast speed while still protecting data privacy due to the post-processing property~\cite{dp}. In contrast, DPSDA incurs the same significant time cost for each synthesis query, as discussed in Section~\ref{subsec:ComputationalResource}. Moreover, their privacy analysis cannot ensure the security of multiple synthesis queries.

For privacy budget 5 and 1, \toolname consistently surpasses all baselines. Particularly, as $\varepsilon$ shifts from 10 to 1, the FID of synthetic {\tt CIFAR-10} images produced by \toolnameD and \toolnameG drops only by 7.8\% and 7.2\%. In contrast, the FID for the state-of-the-art method diminishes by 23.7\%, and for DPGAN-p, it decreases by a substantial 41.0\%. For {\tt CelebA32} and {\tt CelebA64}, \toolnameD and \toolnameG only decreases by 36.1\% and 18.1\% respectively, while the SOTA method decreases by 49.1\% and 50.6\%. However, our \toolnameG decreases by 68.2\% on {\tt CelebA32}. It is well known that GAN suffers from unstable training~\cite{wgan}. Increased gradient noise due to a restricted privacy budget can further intensify this instability. Hence, we believe diffusion models might be more appropriate as image generative models in scenarios with limited privacy budgets.

\begin{figure*}[t]
    \centering
    \setlength{\abovecaptionskip}{0pt}
    \includegraphics[width=0.96\linewidth]{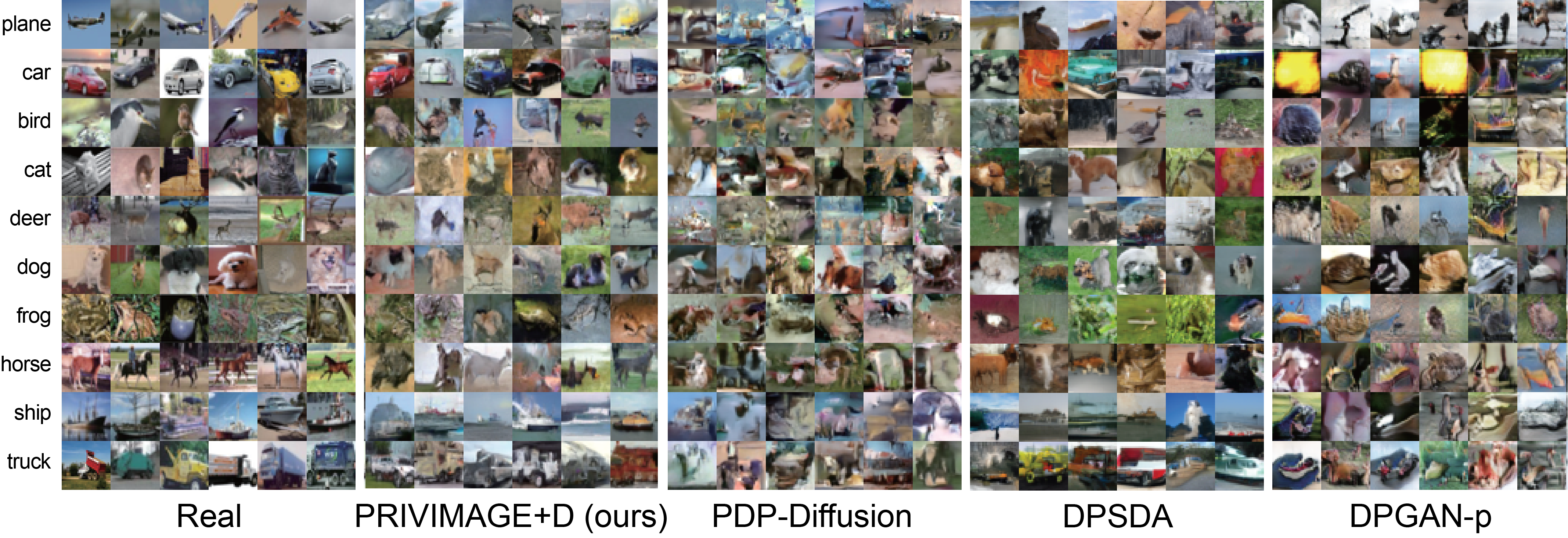}
    \caption{Examples of Synthetic {\tt CIFAR-10}~\cite{cifar10} images with $\varepsilon=10$. These generative models are trained using our tools \toolnameD, as well as PDP-Diffusion~\cite{dpdm-sota} and DPGAN-p. Each row corresponds to a category from the {\tt CIFAR-10} dataset. }
    \label{fig:cifar10_comp_show}
    \vspace{-4.0pt}
\end{figure*}

Additionally, we investigate the distribution characteristics of the synthetic dataset to validate the high utility of our generated data. Specifically, we first use trained CNN classifiers from the dataset generated by \toolnameD, PDP-Diffusion, DPSDA, and DPGAN-p, to undertake the classification task within the {\tt CIFAR-10}. With the aid of the well-trained CNN, each image is transformed into an embedding vector. This vector encapsulates the most pertinent feature information utilized by the CNN for classification. We use t-SNE to visualize them in a two-dimension space. Figure~\ref{fig:tsne} illustrates that embeddings of image data from \toolnameD are projected into 10 clusters aligning with 10 categories. This suggests that the classifier adeptly discerns the unique features of images, thereby achieving superior performance compared to the CNNs trained on the dataset provided by other methods. In contrast, dataset embeddings from others do not form distinct clusters, leading their classifiers to struggle with classification.
Examples of synthetic images for {\tt CIFAR-10} from various methods are presented in Figure~\ref{fig:cifar10_comp_show}. Additionally, we present more synthetic images for {\tt CelebA} and {\tt Camelyon17}~\cite{camelyon1} in Appendix~\ref{ap:moreSamples}.

\begin{table}[h]
\vspace{-8.0pt}
\setlength{\tabcolsep}{3pt}
    \centering
    \renewcommand\arraystretch{1}
    \begin{tabular}{p{0.95\columnwidth}}
    \Xhline{1.0pt}
         \rowcolor{gray0} \noindent \textbf{Answers to RQ1}: Synthetic images produced by \toolname exhibit greater fidelity and utility compared to all baseline methods with three distinct privacy budgets. On average, the FID of the synthetic dataset is \textbf{6.8\%} lower, and the CA of the downstream classification task is \textbf{13.2\%} higher, compared to the state-of-the-art method.\\
    \Xhline{1.0pt}
    \end{tabular}
\vspace{-8.0pt}
\end{table}

% \noindent \textbf{Answers to RQ1}: Synthetic images produced by \toolname exhibit greater fidelity and utility compared to all baseline methods. On average, the FID of the synthetic dataset is \textbf{29.8\%} lower, and the CA of the downstream classification task is \textbf{14.3\%} higher, compared to the state-of-the-art method. 

\subsection*{RQ2. How do the semantic distribution queries of \toolname improve the fine-tuning?}

\begin{figure}[t]
\vspace{+5.0pt}
    \centering
    \includegraphics[width=0.97\linewidth]{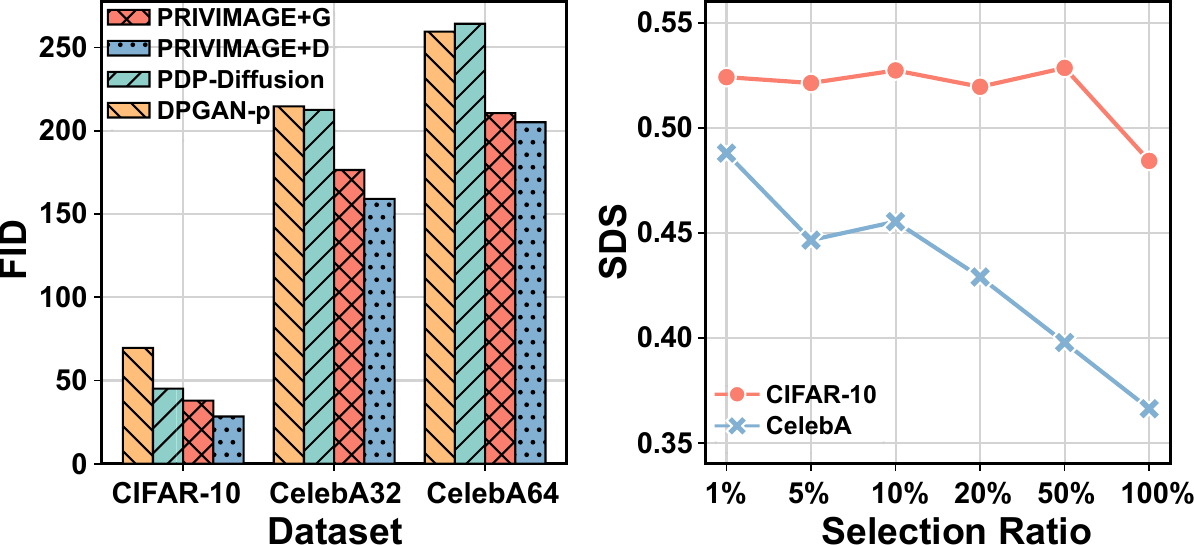}
    \caption{(Left) The FID of baselines pre-trained on the entire public dataset and ours \toolname pre-trained on the selected public dataset.
    (Right) The SDS between the sensitive dataset and different pre-training datasets.}
    \label{fig:rq2}
            \vspace{-4.0pt}
\end{figure}

\noindent \textbf{Experiment Design.} We conducted experiments to explore whether the fine-tuning of generative models is enhanced by leveraging our proposed semantic distribution query to select images from the entire public dataset for pre-training. As a result, \toolname can synthesize images with greater fidelity and utility. We approach this from two perspectives: data distribution similarity and semantic distribution similarity. To evaluate data distribution similarity, we compute the FID of synthetic images generated by different methods, where generative models have only been pre-trained and not fine-tuned on the sensitive dataset. To evaluate semantic distribution similarity, we assess the SDS (as given in Eq.~\eqref{eq:sds}) between the sensitive dataset and the pre-training dataset across various selection ratios. Additionally, we present a showcase from the pre-training dataset selected by our \toolname, to further verify the semantic distribution similarity from a visual perspective in Appendix~\ref{ap:visualSimi}.

\begin{figure*}[!t]
    \centering
    \setlength{\abovecaptionskip}{0pt}
    \includegraphics[width=0.98\linewidth]{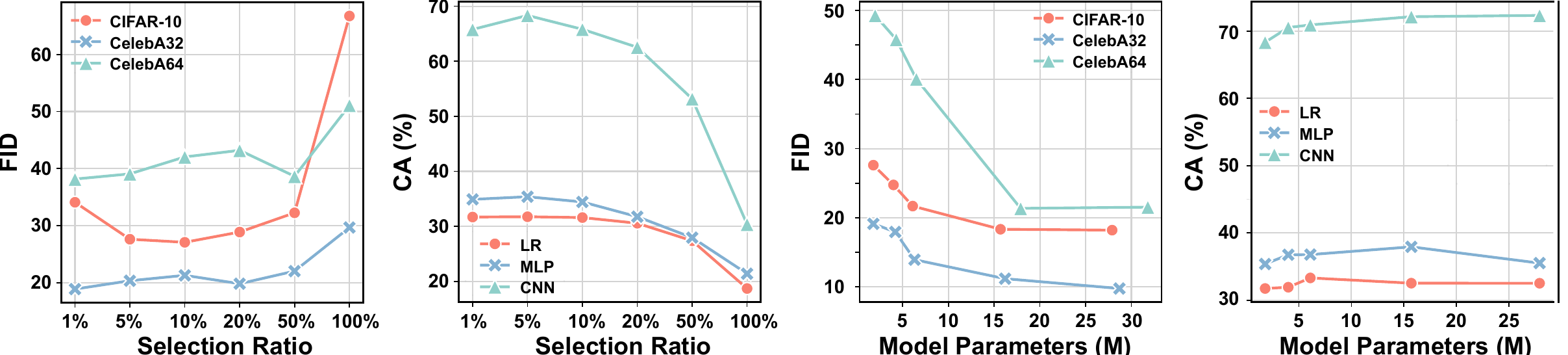}
    \caption{The two left figures present the effect of the selection ratio on the FID of the synthesized image dataset and the CA for downstream classification tasks. The models are trained on datasets selected from the public dataset for pre-training, using various selection ratios. "100\%" represents that pre-training generative models on the entire public dataset. The right two figures display the effect of model size, where the selection ratios are set as 5\% and 1\% for {\tt CIFAR-10} and {\tt CelebA} respectively.}
    \label{fig:rq4}
    \vspace{-4.0pt}
\end{figure*}

\noindent \textbf{Result Analysis.} We analyze our results from two similarity between the synthesized and sensitive dataset as follows. 

\noindent \textit{Data Distribution Similarity.} The left panel of Figure~\ref{fig:rq2} presents that when generative models are pre-trained on the selectively curated public dataset, \toolname produces datasets with superior FID results compared to baselines pre-trained on the entire public dataset. Before fine-tuning on the sensitive dataset, \toolname can intuitively generate synthetic images with a data distribution closer to the sensitive data than images produced by existing methods. \textit{\toolname wins at the beginning.} With an equal number of fine-tuning iterations, \toolname still achieves superior performance.

\noindent \textit{Semantic Distribution Similarity.} For the {\tt CIFAR-10} dataset~\cite{cifar10}, we adopt its 10 category labels as its semantic representation. Similarly, for the {\tt CelebA} dataset~\cite{celeba}, we use its 40 face attribute labels to represent its semantics. The public dataset {\tt ImageNet}~\cite{imagenet} is characterized by its 1000 category labels. We experiment with various selection ratios from this public dataset by adjusting the number of queried semantics, $k_2$, as described in Section~\ref{subsec:sdq}. The right panel of Figure~\ref{fig:rq2} displays that the semantic distribution similarity between {\tt ImageNet}~\cite{imagenet} and {\tt CIFAR-10}\cite{cifar10} surpasses that of {\tt CelebA}\cite{celeba}, aligning with our human visual perceptions as presented in Figure~\ref{fig:dataset_samples}, indicating the utility of our proposed SDS metric. The datasets selected by our \toolname exhibit a closer semantic distribution to the two sensitive datasets than the entire public dataset across different selection ratios. Leveraging the semantic distribution query, \toolname selects a pre-training dataset whose semantic distribution closely mirrors that of the sensitive dataset, all while operating within a minimal privacy budget.

\begin{table}[H]
\vspace{-8.0pt}
\setlength{\tabcolsep}{3pt}
    \centering
    \renewcommand\arraystretch{1}
    \begin{tabular}{p{0.95\columnwidth}}
    \Xhline{1.0pt}
         \rowcolor{gray0} \noindent \textbf{Answers to RQ2}: \toolname selects a pre-training dataset that resembles the sensitive data more closely than the entire public data. Before fine-tuning the sensitive dataset, \toolname produces synthetic images with a data distribution maligned with the sensitive data. As a result, \toolname delivers enhanced DP image synthesis.\\
    \Xhline{1.0pt}
    \end{tabular}
\vspace{-8.0pt}
\end{table}

\subsection*{RQ3. How do hyper-parameters affect the performance of \toolname?}

\noindent \textbf{Experiment Design.} This experiment investigates how the selection ratio of selecting pre-training data from the entire public data and the generative model size impact the performance of \toolnameD. The selection ratio is determined by $k_2$. For example, {\tt ImageNet} has 1,000 subsets, each containing nearly the same number of images. When $k_2=10$, it means that 10 subsets will be selected for pre-training, resulting in a selection ratio of 1\%. We explore selection ratios of {1\%, 5\%, 10\%, 20\%, 50\%}. In terms of model size, we consider {1.8M, 4.0M, 6.1M, 15.7M, 27.9M} for {\tt CIFAR-10}, {1.8M, 4.2M, 6.3M, 16.2M, 28.7M} for {\tt CelebA32}, and {2.0M, 4.3M, 6.5M, 17.9M, 31.8M} for {\tt CelebA64}.

\noindent \textbf{Result Analysis.} Figure~\ref{fig:rq4} shows that across all selection ratios, \toolname outperforms the full public dataset (where the selection ratio is 100\%), revealing the effectiveness of our semantic distribution query. As the selection ratio increases, implying more data is used for pre-training, \toolname generally exhibits poorer FID and CA on both {\tt CIFAR-10} and {\tt CelebA}. We believe the reason for this is that only a small portion of {\tt ImageNet} is similar to our sensitive datasets. Pre-training models on a larger number of dissimilar images may distract the models from learning the distribution of similar images, making it more challenging for the models to learn the distribution of the sensitive datasets. However, this trend is more pronounced for {\tt CIFAR-10} than for {\tt CelebA}. This can be attributed to the fact that {\tt CelebA} diverges more from the public dataset {\tt ImageNet}, resulting in fewer opportunities for \toolname to select beneficial data for pre-training. For {\tt CIFAR-10}, the performance of \toolname is notably impacted. Specifically, when the selection ratio is at 1\%, the FID of synthetic images rises compared to that at 5\%. Hence, when there is a significant discrepancy between sensitive and public data, we have the flexibility to select a ratio, which typically outperforms using the entire public dataset. To further investigate this phenomenon, we conduct experiments on another dataset with a larger domain shift from {\tt ImageNet}. The results of these experiments are presented in Appendix~\ref{ap:camelyon17}.

\begin{table}[!t]
\vspace{-5.0pt}
\setlength{\tabcolsep}{3pt}
\small
    \centering
    \caption{The FID and CA of \toolname and three non-private baselines on {\tt CIFAR-10}~\cite{cifar10}, {\tt CelebA32} and {\tt CelebA64}~\cite{celeba} with $\varepsilon=10$.}
    \label{tab:dis1}
    \begin{tabular}{l|cccc|c|c}
    \toprule
    \multirow{3}{*}{Method} & \multicolumn{4}{c|}{{\tt CIFAR-10}} & {\tt CelebA32} & {\tt CelebA64 }\\
    \Xcline{2-7}{0.5pt}
    & \multicolumn{3}{c}{CA (\%)} & \multirow{2}{*}{FID} & \multirow{2}{*}{FID} & \multirow{2}{*}{FID}\\
     \Xcline{2-4}{0.5pt}
     & LR & MLP & CNN & &\\
    \hline
0    NonPrivG & 18.2 & 24.7 & 49.8 & 22.5 & 13.6 & 34.7\\
    NonPrivD & 35.8 & 42.2 & 77.1  & 19.8 & 9.01 & 18.0\\
    NonPriv & 37.4 & 45.7 & 86.1 & - & - & -\\
    \hline
    \rowcolor{gray0} \toolnameG & 19.9 & 24.5 & 44.3 & 28.1 & 18.9 & 38.2\\
    \rowcolor{gray0} \toolnameD & 31.7 & 35.4 & 68.3 & 27.6 & 19.1 & 49.3\\
    \bottomrule
\end{tabular}
\vspace{-8.0pt}
\end{table}
In terms of model size, as it increases, \toolname generally achieves superior FID and CA results. However, as the model size becomes significantly larger, the performance gets unstable. For instance, the CA of MLP classifiers trained on synthetic images from \toolname at 27.9M is lower than at 15.7M. Similarly, the CA of LR classifiers from \toolname at 15.7M is below that of 6.1M. These observations suggest that lighter generative models might be better suited for the DP image synthesis task.

\begin{table}[h]
\vspace{-8.0pt}
\setlength{\tabcolsep}{3pt}
    \centering
    \renewcommand\arraystretch{1}
    \begin{tabular}{p{0.95\columnwidth}}
    \Xhline{1.0pt}
         \rowcolor{gray0} \noindent \textbf{Answers to RQ3}: Compared to using the entire public dataset, \toolname achieves superior performance with a selection ratio less than 50\%. As the divergence between public and sensitive data widens, the influence of the selection ratio diminishes gradually. When the generative model comprises fewer than 15M parameters, \toolname produces higher-quality synthesized images with larger models. Whereas, as the parameters increase further, \toolname experiences unstable training.\\
    \Xhline{1.0pt}
    \end{tabular}
\vspace{-8.0pt}
\end{table}

% \noindent \textbf{Answers to RQ4}: Compared to using the entire public dataset, \toolname achieves superior performance with a selection ratio less than 50\%. As the divergence between public and sensitive data widens, the influence of the selection ratio diminishes gradually. Additionally, when the generative model comprises fewer than 15M parameters, \toolname produces higher-quality synthesized images with larger models.

\section{Discussion}
\label{sec:dis}
This section discusses how \toolname performs without privacy-protective settings, the computational resources required by \toolname, as well as its potential applications and inherent limitations.
We also discuss about how to choose the privacy parameter in Appendix~\ref{ap:epsChoice}.

\subsection{\toolname without Privacy Protection}

This experiment studies how the synthetic performance of \toolname is harmed by adhering to the DP framework. We compared \toolname with three methods: (1) ``NonPrivD'' and ``NonPrivG'' train diffusion models and GANs respectively on the pre-training dataset chosen by \toolname with selection radio 5\%, whereas the gradient in fine-tuning stage and semantic distribution in the pre-training stage are not introduced Gaussian noise. (2) ``NonPriv'' directly uses the sensitive dataset to train the classifier for the downstream classification task. It is noticed that the FID measures the quality of images created by the generative models, so that sensitive datasets do not have an FID value. We conduct these experiments with a privacy budget ($10,1 \times 10^{-5}$).  

Table~\ref{tab:dis1} shows that, on average, \toolnameD and \toolnameG only decrease by 6.5\% and 1.4\% in terms of CA on three classification models. Therefore, with sensitive data protected, \toolname still generates useful synthetic images. In terms of CA, \toolname shows an 11.2\% decrease compared to NonPriv. However, when we increase the generative model size of \toolnameD by 2.2M parameters, the decrease is just 7.9\%. This suggests that more efforts are needed to develop our \toolname further.

\begin{table}[!t]
\vspace{-6.0pt}
\setlength{\tabcolsep}{3pt}
\small
    \centering
    \caption{GPU memory cost, Running time, CA and FID of synthetic images of PDP-Diffusion, DPSDA and \toolname. The time cost of SQF includes the time taken to train SQF and to use the trained SQF for querying the semantic distribution.}
    \begin{tabular}{l|c|c|c|c}
    \toprule
      \multicolumn{2}{c|}{Evaluation Metrics}& PDP-Diffusion & DPSDA & \toolname\\
    \hline
    \multirow{4}{*}{Memory} & Pre-train &  107GB & 0GB & 55GB \\
     & Fine-tune & 158GB & 0GB & 89GB\\
     & SQF & 0GB & 0GB & 22GB\\
     & Synthesis & 177GB & 219GB & 83GB\\
     \hline
    \multirow{4}{*}{Time} & Pre-train  & 87h & 0h & 46h\\ %-66\%
     & Fine-tune  & 24h & 0h & 3h\\ %-99\%
     & SQF  & 0h & 0h & 9h\\
     & Synthesis & 0.37h & 12h & 0.15h\\
     \hline
     \multirow{3}{*}{CA (\%)} & LR & 15.07 & 14.3 & 33.11 \\
     & MLP & 14.17 & 13.4 & 36.78\\
     & CNN & 24.27 & 15.1 & 70.91\\
    \hline
    \multicolumn{2}{c|}{FID} & 22.90 & 16.2 & 21.69\\
    \bottomrule
\end{tabular}
\label{tab:computationalResource}
\vspace{-8.0pt}
\end{table}

\subsection{Consumption of Computational Resource}
\label{subsec:ComputationalResource}

% \begin{table}[!t]
% \vspace{-6.0pt}
% \setlength{\tabcolsep}{3pt}
% \small
%     \centering
%     \caption{GPU memory cost, Running time, CA and FID of synthetic images of PDP-Diffusion and \toolname. `Changes' represents the relative percentage variation in the evaluation metrics of \toolname when compared to PDP-Diffusion. }
%     \begin{tabular}{l|c|c|c|c}
%     \toprule
%       \multicolumn{2}{c|}{Evaluation Metrics}& PDP-Diffusion & \toolname & Changes \\
%     \hline
%     \multirow{4}{*}{Memory} & Pre-train &  107GB & 55GB & \multirow{4}{*}{\textbf{-50\%}}\\
%      & Fine-tune & 158GB & 89GB &\\
%      & SQF & 0 & 22GB &\\
%      & Synthesis & 177GB & 83GB &\\
%      \hline
%     \multirow{4}{*}{Time} & Pre-train  & 87h& 46h & \multirow{4}{*}{\textbf{-48\%}}\\ %-66\%
%      & Fine-tune  & 24h& 3h &\\ %-99\%
%      & SQF  & 0 & 9h &\\
%      & Synthesis & 22min & 9min &\\
%      \hline
%      \multirow{3}{*}{CA (\%)} & LR & 15.07 & 33.11 & \textbf{+18.0}\\
%      & MLP & 14.17 & 36.78 & \textbf{+22.6}\\
%      & CNN & 24.27 & 70.91 & \textbf{+46.6}\\
%     \hline
%     \multicolumn{2}{c|}{FID} & 22.90 & 21.69 & \textbf{-5\%} \\
%     \bottomrule
% \end{tabular}
% \label{tab:computationalResource}
% \vspace{-8.0pt}
% \end{table}

This section presents the high efficiency of \toolname and the significant trade-off between the computation resource cost and synthesis performance. We trained the state-of-the-art method PDP-Diffusion using an 80M diffusion models~\cite{dpdm-sota} and \toolnameD using a 6.1M diffusion model on {\tt CIFAR-10}. We also implement DPSDA~\cite{api} using a 270M diffusion model pre-trained on {\tt ImageNet}. We change their hyper-parameters \{lookahead degree, PE iterations, timestep respacing\} from \{8, 21, 100\} into \{2, 11, 50\} for faster generation, but the performance could be worse than what they reported in their paper. We compare their GPU memory usage and time cost during pre-training, fine-tuning and synthesis on identical computational configurations. \toolname needs additional time to train the SQF.

Table~\ref{tab:computationalResource} shows that \toolname uses only 50\% and 59\% of the GPU memory used by PDP-Diffusion and DPSDA, respectively. In terms of runtime, \toolname is 48\% faster than PDP-Diffusion. These results can be attributed to the compact dataset selected by \toolname for pre-training and lighter generative models that \toolname uses. Although \toolname requires more time compared to DPSDA, the majority of \toolname's time is spent on training, while the synthesis process is 98\% faster than DPSDA. When our synthesis method needs to respond to multiple queries, DPSDA could be significantly slower than \toolname.
% Pre-training PDP-Diffusion on the entire public dataset requires a diffusion model with 80M parameters, while \toolname only uses 5\% of the public dataset to pre-train lightly-parameterized diffusion model with only 6.1M parameters. 
Despite the significant savings in computational resources, our \toolname still achieves an FID and average CA that are 5\% lower and 29\% higher than PDP-Diffusion.

As presented in Table~\ref{tab:rq1}, compared to an 1.8M model, PDP-Diffusion with an 80M model does achieve better FID. However, the average CA on three classifiers drops by 5.7\%, suggesting that the synthetic images are less useful. This is attributed to the fact that the $l_2$-norm of the noise added during DP-SGD scales linearly to the dimension of parameters, leading to a ``curse of dimensionality'' and making the training of generative models unstable~\cite{dpdm,nolargemodel1,nolargemodel2}. We also conduct another experiment to prove this ``curse of dimensionality'' and put the result in Appendix~\ref{ap:OMI}. Besides, we may need to upload our sensitive data to a remote server using such over-parameterized models, presenting potential security concerns~\cite{DPPublicPretraining1}. Therefore, we should be more concerned about achieving great synthesis with lightly parameterized models, which are less affected by DP-SGD and can be used on end-user devices more easily.

% \kc{for the fair comparison, we train all Diffusion-based methods (e.g., DPDM, PDP-Diffusion, and \toolnameD) using the same diffusion model with 1.8M parameters. Using such light models, PDP-Diffusion cannot generate synthetic images with great utility and fidelity for large pre-training dataset they use. In this paper, we more concern about how to achieve great synthesis with lightly-parameterized models, which are less affected by DP-SGD and can be employed on end-user devices more easily. It is expensive to employ PDP-Diffusion with 80M models as they proposes. Besides, even though PDP-Diffusion uses diffusion models with 80M parameters, our \toolnameD can still achieve competitive or even better results. For example, \toolnameD achieves 7.6\% and 19.1\% lower FID with 4M and 6.1M parameters respectively. Therefore, it is not meaningful to choose PDP-Diffusion with 80M models as our baselines.}

\subsection{Applications and Limitations}

This subsection discusses the application scope and limitations of \toolname.

\noindent \textbf{Potential Applicability to Other Fields.} Though this paper primarily focuses on image data synthesis, we posit that \toolname has high potential applicability to other data types, such as text data. Numerous studies have highlighted the effectiveness of diffusion models in tasks like text generation~\cite{diffusionText1} and audio synthesis~\cite{diffusionAudio1}. Analogous to the practical adopted in \toolname, by judiciously selecting a suitable public dataset as well as integrating pre-training with fine-tuning methodologies, it seems feasible to achieve DP text, audio synthesis, and even other fields.

\noindent \textbf{Dependence on Public Data.} The primary advantage of \toolname lies in its ability to query the semantic distribution of sensitive data and subsequently select data from a public dataset that closely aligns with the sensitive dataset for pre-training. However, when the distribution of the public data substantially deviates from that of the sensitive data, \toolname faces limited opportunities to pick out such pertinent data. In the absence of a suitable public dataset, the effectiveness of \toolname in generating complex image datasets while adhering to the DP guarantee diminishes.

% \subsection{Real-World Application Scenarios}

% \toolname can produce synthetic images of exceptional fidelity and utility using compact generative models with just \textbf{1.8M} parameters. Furthermore, it leverages a minimal pre-training dataset consisting of merely \textbf{13K} images. Such efficiency and compactness make \toolname highly suitable for deployment on a plethora of end-user devices. This obviates the necessity to transfer sensitive data to remote cloud services, enhancing user privacy and data security. \toolname lets organizations use and share synthetic images without privacy worries, aiding the creation of strong machine learning models without compromising data privacy.
% \tw{feel this can be deleted}
% \tw{you can put some limitations here, e.g., we assume sensitive dataset must be within imagenet and cannot handle strange datasets}

\section{Related Work}
\label{sec:related_work}

This section explicitly discusses two main types of DP image data synthesis works, including sanitizing generative models through DP-SGD~\cite{dpsgd} and querying DP information of sensitive datasets for synthesis.

\noindent \textbf{Sanitizing Generative Models via DP-SGD.}
Currently, most efforts have focused on applying DP-SGD~\cite{dpsgd} on popular generative models, like GANs~\cite{dpsgd-gan1,dpsgd-gan2,dpsgd-gan4,dpsgd-gan6}, diffusion models~\cite{dpdm,dpldm,dpdm-sota} or variational autoencoders (VAE)~\cite{dp2vae,dps-fvae,lvae}. They introduce noise into gradient calculations during training, ensuring that individual data points do not disproportionately influence the model's learning, thereby offering a degree of privacy protection. 
Diffusion models have shown promise in generating DP image data. For example, PDP-Diffusion~\cite{dpdm-sota} and DPDM~\cite{dpdm} respectively achieve state-of-the-art performance in terms of fidelity and utility of synthetic images, and demonstrate top-tier classification accuracy, both under the setting of with and without the use of a public dataset for pre-training. However, DPDM only achieves synthesis on some naive datasets (e.g. {\tt MNIST}). PDP-Diffusion demands significant computational resources and struggles to produce images with great fidelity and utility on end-user devices for real-world applications~\cite{DPPublicPretraining1}.

\toolname presented in this paper also relies on sanitizing generative models via DP-SGD for DP image synthesis. Different from existing methods, \toolname queries the semantic distribution of the sensitive dataset to select a more compact dataset from the public dataset for pre-training. \toolname enables us to use a lightly parameterized model, while achieving superior synthesis performance.

\noindent \textbf{Querying DP Information for Synthesis.}
Another type of method emphasizes designing a query function to extract valuable features from a sensitive dataset~\cite{dp-mepf,dp-merf,pearl,api}. These features exhibit low sensitivity and aid in obtaining pertinent information for synthesis. 
% We query the marginal of tabular dataset whose sensitivity is 1~\cite{privsyn}. When incorporating Gaussian noise, the noisy features can be viewed as a target. We simply need to devise a method (e.g., GANs) to produce a synthetic dataset with features closely resembling this target~\cite{dp-merf,pearl,dp-mepf}. 
We will next introduce the various feature query functions without delving into the generation process.

Harder et al. presented DP-MERF~\cite{dp-merf}, a method that represents the sensitive dataset using random features like Fourier features. Meanwhile, Seng et al.~\cite{pearl} suggested substituting the random feature with a characteristic function, which offers improved generalization capability. Harder et al. introduced DP-MEPF~\cite{dp-mepf}, a method that leverages public data to extract perceptual features. Specifically, they represent each image as a feature vector using an encoder pre-trained on public datasets, such as {\tt ImageNet}~\cite{imagenet}. These methods, however, fall short in comparison to DP-SGD~\cite{dpsgd} in term of generated images qualification, especially for more colorful dataset (e.g., {\tt CIFAR-10}~\cite{cifar10}). Recently, DPSDA~\cite{api} proposed to utilize foundation model APIs of powerful generative models trained on public datasets, such as DALLE-2~\cite{dalle2}, Stable Diffusion~\cite{StableDiffusion}, and GPT3/4~\cite{gpt3,gpt4}. Although this method achieves results comparable to the SOTA~\cite{dpdm-sota}, DPSDA heavily depends on two open-source APIs, RANDOM-API and VARIATION-API, which may not be available in some scenarios, especially for the VARIATION-API.

\section{Conclusions and Further Works}
\label{sec:conclusions}
This paper explores leveraging the public dataset more effectively to pre-train lightly parameterized models for DP image synthesis.
We propose \toolname for generating a synthetic image dataset under differential privacy. Compared to existing methods, \toolname queries the semantic distribution of sensitive data to select more useful data for pre-training. With lightly parameterized generative models and a small pre-training dataset, \toolname still generates synthetic images with excellent fidelity and utility. Besides, \toolname saves much computational resource compared to the state-of-the-art method DP-Diffusion and can be employed on end-user devices more easily. This paper calls for attention to construct a more tailored pre-training dataset to advance the practical implementations of DP image dataset synthesis.

Future work plans to address the challenge of generating high-quality images when the sensitive dataset greatly diverges from the public dataset. We aspire to extend the \toolname to synthesize other types of data and develop \toolname into a more practical DP image synthesis tool.

\section*{Acknowledgement}
We thank all the anonymous reviewers and our shepherd for their valuable comments. Authors from CAS in this 
 research/project are supported by the National Science and Technology Project (2022ZD0116406). Kecen Li’s work was done as a remote intern at UVA.

% \clearpage
\bibliographystyle{plain}
\bibliography{sample}

\begin{thebibliography}{10}

\bibitem{dpsgd}
Mart{\'{\i}}n Abadi, Andy Chu, Ian~J. Goodfellow, and et~al.
\newblock Deep learning with differential privacy.
\newblock In {\em Proceedings of the 2016 {ACM} {SIGSAC} Conference on Computer
  and Communications Security}, pages 308--318, 2016.

\bibitem{wgan}
Mart{\'{\i}}n Arjovsky, Soumith Chintala, and L{\'{e}}on Bottou.
\newblock Wasserstein generative adversarial networks.
\newblock In {\em Proceedings of the 34th International Conference on Machine
  Learning, {ICML}}, volume~70 of {\em Proceedings of Machine Learning
  Research}, pages 214--223, 2017.

\bibitem{dpsgd-gan1}
Sean Augenstein, H.~Brendan McMahan, Daniel Ramage, and et~al.
\newblock Generative models for effective {ML} on private, decentralized
  datasets.
\newblock {\em CoRR}, abs/1911.06679, 2019.

\bibitem{camelyon1}
P{\'{e}}ter B{\'{a}}ndi, Oscar Geessink, Quirine Manson, and et~al.
\newblock From detection of individual metastases to classification of lymph
  node status at the patient level: The {CAMELYON17} challenge.
\newblock {\em {IEEE} Trans. Medical Imaging}, 38(2):550--560, 2019.

\bibitem{CVPretaining2}
Hangbo Bao, Li~Dong, Songhao Piao, and et~al.
\newblock Beit: {BERT} pre-training of image transformers.
\newblock In {\em The Tenth International Conference on Learning
  Representations, {ICLR}}, 2022.

\bibitem{DBLP:journals/corr/abs-2112-02918}
Franziska Boenisch, Adam Dziedzic, Roei Schuster, and et~al.
\newblock When the curious abandon honesty: Federated learning is not private.
\newblock {\em CoRR}, abs/2112.02918, 2021.

\bibitem{biggan}
Andrew Brock, Jeff Donahue, and Karen Simonyan.
\newblock Large scale {GAN} training for high fidelity natural image synthesis.
\newblock In {\em 7th International Conference on Learning Representations,
  {ICLR}}, 2019.

\bibitem{gpt3}
Tom~B. Brown, Benjamin Mann, Nick Ryder, and et~al.
\newblock Language models are few-shot learners.
\newblock In {\em Advances in Neural Information Processing Systems}, 2020.

\bibitem{dpgenreview}
Dingfan Chen, Raouf Kerkouche, and Mario Fritz.
\newblock A unified view of differentially private deep generative modeling.
\newblock {\em CoRR}, abs/2309.15696, 2023.

\bibitem{dpsgd-gan2}
Dingfan Chen, Tribhuvanesh Orekondy, and Mario Fritz.
\newblock {GS-WGAN:} {A} gradient-sanitized approach for learning
  differentially private generators.
\newblock In {\em Advances in Neural Information Processing Systems}, 2020.

\bibitem{dpsgd-gan3}
Dongjie Chen, Sen{-}ching~Samson Cheung, Chen{-}Nee Chuah, and et~al.
\newblock Differentially private generative adversarial networks with model
  inversion.
\newblock In {\em {IEEE} International Workshop on Information Forensics and
  Security, {WIFS}}, pages 1--6, 2021.

\bibitem{COCOCaption}
Xinlei Chen, Hao Fang, Tsung{-}Yi Lin, and et~al.
\newblock Microsoft {COCO} captions: Data collection and evaluation server.
\newblock {\em CoRR}, abs/1504.00325, 2015.

\bibitem{DBLP:journals/tdp/DankarE13}
Fida~Kamal Dankar and Khaled~El Emam.
\newblock Practicing differential privacy in health care: {A} review.
\newblock {\em Trans. Data Priv.}, 6(1):35--67, 2013.

\bibitem{imagenet}
Jia Deng, Wei Dong, Richard Socher, and et~al.
\newblock Imagenet: {A} large-scale hierarchical image database.
\newblock In {\em {IEEE} Computer Society Conference on Computer Vision and
  Pattern Recognition {CVPR}}, pages 248--255, 2009.

\bibitem{dpdm}
Tim Dockhorn, Tianshi Cao, Arash Vahdat, and et~al.
\newblock Differentially private diffusion models.
\newblock {\em CoRR}, 2022.

\bibitem{DBLP:journals/corr/abs-2104-01987}
Jinshuo Dong, Aaron Roth, and Weijie~J. Su.
\newblock Rejoinder: Gaussian differential privacy.
\newblock {\em CoRR}, abs/2104.01987, 2021.

\bibitem{dp}
Cynthia Dwork, Frank McSherry, Kobbi Nissim, and Adam~D. Smith.
\newblock Calibrating noise to sensitivity in private data analysis.
\newblock In {\em {TCC}}, pages 265--284, 2006.

\bibitem{dpdm-sota}
Sahra Ghalebikesabi, Leonard Berrada, Sven Gowal, and et~al.
\newblock Differentially private diffusion models generate useful synthetic
  images.
\newblock {\em CoRR}, 2023.

\bibitem{diffusionAudio1}
Karan Goel, Albert Gu, Chris Donahue, and et~al.
\newblock It's raw! audio generation with state-space models.
\newblock In {\em International Conference on Machine Learning, {ICML}}, volume
  162 of {\em Proceedings of Machine Learning Research}, pages 7616--7633,
  2022.

\bibitem{10.1145/3564625.3564636}
Chen Gong, Zhou Yang, Yunpeng Bai, and et~al.
\newblock Curiosity-driven and victim-aware adversarial policies.
\newblock In {\em Proceedings of the 38th Annual Computer Security Applications
  Conference}, ACSAC '22, page 186–200, 2022.

\bibitem{GONG2020107310}
Chen Gong, Xiaoxiong Zhang, and Yunyun Niu.
\newblock Identification of epilepsy from intracranial eeg signals by using
  different neural network models.
\newblock {\em Computational Biology and Chemistry}, 87:107310, 2020.

\bibitem{gong2022pattern}
Chen Gong, Xingchen Zhou, and Yunyun Niu.
\newblock Pattern recognition of epilepsy using parallel probabilistic neural
  network.
\newblock {\em Applied Intelligence}, 52(2):2001--2012, 2022.

\bibitem{gan}
Ian~J. Goodfellow, Jean Pouget{-}Abadie, Mehdi Mirza, and et~al.
\newblock Generative adversarial nets.
\newblock In {\em Advances in Neural Information Processing Systems}, pages
  2672--2680, 2014.

\bibitem{lvae}
Benedikt Gro{\ss} and Gerhard Wunder.
\newblock Differentially private synthetic data generation via
  lipschitz-regularised variational autoencoders.
\newblock {\em CoRR}, abs/2304.11336, 2023.

\bibitem{dp-merf}
Frederik Harder, Kamil Adamczewski, and Mijung Park.
\newblock {DP-MERF:} differentially private mean embeddings with randomfeatures
  for practical privacy-preserving data generation.
\newblock In {\em {AISTATS}}, volume 130, pages 1819--1827, 2021.

\bibitem{dp-mepf}
Frederik Harder, Milad Jalali, Danica~J. Sutherland, and et~al.
\newblock Pre-trained perceptual features improve differentially private image
  generation.
\newblock {\em Trans. Mach. Learn. Res.}, 2023, 2023.

\bibitem{DBLP:journals/jpdc/HassanRC20}
Muneeb~Ul Hassan, Mubashir~Husain Rehmani, and Jinjun Chen.
\newblock Differential privacy in blockchain technology: {A} futuristic
  approach.
\newblock {\em J. Parallel Distributed Comput.}, 145:50--74, 2020.

\bibitem{resnet}
Kaiming He, Xiangyu Zhang, Shaoqing Ren, and et~al.
\newblock Deep residual learning for image recognition.
\newblock In {\em 2016 {IEEE} Conference on Computer Vision and Pattern
  Recognition, {CVPR}}, pages 770--778, 2016.

\bibitem{DPPublicPretraining3}
Dan Hendrycks, Collin Burns, Anya Chen, and et~al.
\newblock {CUAD:} an expert-annotated {NLP} dataset for legal contract review.
\newblock In {\em Proceedings of the Neural Information Processing Systems
  Track on Datasets and Benchmarks 1}, 2021.

\bibitem{scalinglaw2}
Tom Henighan, Jared Kaplan, and et~al.
\newblock Scaling laws for autoregressive generative modeling.
\newblock {\em CoRR}, abs/2010.14701, 2020.

\bibitem{modelDataSize}
Joel Hestness, Sharan Narang, Newsha Ardalani, and et~al.
\newblock Deep learning scaling is predictable, empirically.
\newblock {\em CoRR}, abs/1712.00409, 2017.

\bibitem{ddpm}
Jonathan Ho, Ajay Jain, and Pieter Abbeel.
\newblock Denoising diffusion probabilistic models.
\newblock In {\em Advances in Neural Information Processing Systems}, 2020.

\bibitem{dppoison2}
Sanghyun Hong, Varun Chandrasekaran, and et~al.
\newblock On the effectiveness of mitigating data poisoning attacks with
  gradient shaping.
\newblock {\em arXiv preprint arXiv:2002.11497}, 2020.

\bibitem{dp2vae}
Dihong Jiang, Guojun Zhang, Mahdi Karami, and et~al.
\newblock Dp\({}^{\mbox{2}}\)-vae: Differentially private pre-trained
  variational autoencoders.
\newblock {\em CoRR}, abs/2208.03409, 2022.

\bibitem{DBLP:journals/corr/abs-2010-02973}
Honglu Jiang, Jian Pei, Dongxiao Yu, and et~al.
\newblock Differential privacy and its applications in social network analysis:
  {A} survey.
\newblock {\em CoRR}, abs/2010.02973, 2020.

\bibitem{dpsgd-gan4}
James Jordon, Jinsung Yoon, and Mihaela van~der Schaar.
\newblock {PATE-GAN:} generating synthetic data with differential privacy
  guarantees.
\newblock In {\em 7th International Conference on Learning Representations,
  {ICLR}}, 2019.

\bibitem{scalinglaw1}
Jared Kaplan, Sam McCandlish, and et~al.
\newblock Scaling laws for neural language models.
\newblock {\em CoRR}, abs/2001.08361, 2020.

\bibitem{stylegan}
Tero Karras, Samuli Laine, and Timo Aila.
\newblock A style-based generator architecture for generative adversarial
  networks.
\newblock In {\em {IEEE} Conference on Computer Vision and Pattern Recognition,
  {CVPR}}, pages 4401--4410, 2019.

\bibitem{stylegan2}
Tero Karras, Samuli Laine, Miika Aittala, and et~al.
\newblock Analyzing and improving the image quality of stylegan.
\newblock In {\em {IEEE/CVF} Conference on Computer Vision and Pattern
  Recognition, {CVPR}}, pages 8107--8116, 2020.

\bibitem{camelyon2}
Pang~Wei Koh, Shiori Sagawa, Henrik Marklund, and et~al.
\newblock {WILDS:} {A} benchmark of in-the-wild distribution shifts.
\newblock In {\em Proceedings of the 38th International Conference on Machine
  Learning, {ICML}}, volume 139 of {\em Proceedings of Machine Learning
  Research}, pages 5637--5664, 2021.

\bibitem{cifar10}
Alex Krizhevsky, Geoffrey Hinton, et~al.
\newblock Learning multiple layers of features from tiny images.
\newblock 2009.

\bibitem{nolargemodel1}
Alexey Kurakin, Steve Chien, Shuang Song, and et~al.
\newblock Toward training at imagenet scale with differential privacy.
\newblock {\em CoRR}, abs/2201.12328, 2022.

\bibitem{ffcv}
Guillaume Leclerc, Andrew Ilyas, Logan Engstrom, and et~al.
\newblock {FFCV}: Accelerating training by removing data bottlenecks.
\newblock In {\em Computer Vision and Pattern Recognition (CVPR)}, 2023.

\bibitem{mnist}
Yann LeCun, L{\'{e}}on Bottou, Yoshua Bengio, and et~al.
\newblock Gradient-based learning applied to document recognition.
\newblock {\em Proc. {IEEE}}, 86(11):2278--2324, 1998.

\bibitem{diffusionText1}
Xiang Li, John Thickstun, Ishaan Gulrajani, and et~al.
\newblock Diffusion-lm improves controllable text generation.
\newblock In {\em NeurIPS}, 2022.

\bibitem{CVPretaining3}
Xiujun Li, Xi~Yin, Chunyuan Li, and et~al.
\newblock Oscar: Object-semantics aligned pre-training for vision-language
  tasks.
\newblock In {\em ECCV}, pages 121--137. Springer, 2020.

\bibitem{AwareDP3}
Xuechen Li, Florian Tram{\`{e}}r, Percy Liang, and et~al.
\newblock Large language models can be strong differentially private learners.
\newblock In {\em The Tenth International Conference on Learning
  Representations, {ICLR}}, 2022.

\bibitem{pearl}
Seng~Pei Liew, Tsubasa Takahashi, and Michihiko Ueno.
\newblock {PEARL:} data synthesis via private embeddings and adversarial
  reconstruction learning.
\newblock In {\em The Tenth International Conference on Learning
  Representations, {ICLR}}, 2022.

\bibitem{COCODetection}
Tsung{-}Yi Lin, Michael Maire, Serge~J. Belongie, and et~al.
\newblock Microsoft {COCO:} common objects in context.
\newblock In {\em ECCV}, volume 8693 of {\em Lecture Notes in Computer
  Science}, pages 740--755, 2014.

\bibitem{api}
Zinan Lin, Sivakanth Gopi, Janardhan Kulkarni, and et~al.
\newblock Differentially private synthetic data via foundation model apis 1:
  Images.
\newblock {\em CoRR}, abs/2305.15560, 2023.

\bibitem{Semantic2}
Huan Ling, Karsten Kreis, Daiqing Li, and et~al.
\newblock Editgan: High-precision semantic image editing.
\newblock In {\em Advances in Neural Information Processing Systems}, pages
  16331--16345, 2021.

\bibitem{dpsgd-gan6}
Yi~Liu, Jialiang Peng, James Jian~Qiao Yu, and et~al.
\newblock {PPGAN:} privacy-preserving generative adversarial network.
\newblock In {\em 25th {IEEE} International Conference on Parallel and
  Distributed Systems, {ICPADS}}, pages 985--989, 2019.

\bibitem{celeba}
Ziwei Liu, Ping Luo, Xiaogang Wang, and et~al.
\newblock Deep learning face attributes in the wild.
\newblock In {\em 2015 {IEEE} International Conference on Computer Vision,
  {ICCV} 2015,}, pages 3730--3738, 2015.

\bibitem{Semantic1}
Jia Long and Hongtao Lu.
\newblock Multi-level gate feature aggregation with spatially adaptive
  batch-instance normalization for semantic image synthesis.
\newblock In {\em MultiMedia Modeling - 27th International Conference, {MMM}},
  volume 12572, pages 378--390, 2021.

\bibitem{dpldm}
Saiyue Lyu, Margarita Vinaroz, Michael~F. Liu, and et~al.
\newblock Differentially private latent diffusion models.
\newblock {\em CoRR}, 2023.

\bibitem{dppoison1}
Yuzhe Ma, Xiaojin Zhu, and Justin Hsu.
\newblock Data poisoning against differentially-private learners: Attacks and
  defenses.
\newblock In {\em Proceedings of the Twenty-Eighth International Joint
  Conference on Artificial Intelligence, {IJCAI}}, pages 4732--4738, 2019.

\bibitem{CrossEntropy}
Anqi Mao, Mehryar Mohri, and Yutao Zhong.
\newblock Cross-entropy loss functions: Theoretical analysis and applications.
\newblock In {\em International Conference on Machine Learning, {ICML}}, volume
  202 of {\em Proceedings of Machine Learning Research}, pages 23803--23828,
  2023.

\bibitem{diffusionAttacker3}
Tomoya Matsumoto, Takayuki Miura, and Naoto Yanai.
\newblock Membership inference attacks against diffusion models.
\newblock In {\em 2023 {IEEE} Security and Privacy Workshops (SPW)}, pages
  77--83.

\bibitem{rdp}
Ilya Mironov.
\newblock Renyi differential privacy.
\newblock {\em CoRR}, abs/1702.07476, 2017.

\bibitem{sgm}
Ilya Mironov, Kunal Talwar, and Li~Zhang.
\newblock R{\'{e}}nyi differential privacy of the sampled gaussian mechanism.
\newblock {\em CoRR}, abs/1908.10530, 2019.

\bibitem{iddpm}
Alexander~Quinn Nichol and Prafulla Dhariwal.
\newblock Improved denoising diffusion probabilistic models.
\newblock In {\em Proceedings of the 38th International Conference on Machine
  Learning, {ICML}}, pages 8162--8171, 2021.

\bibitem{gpt4}
OpenAI.
\newblock {GPT-4} technical report.
\newblock {\em CoRR}, abs/2303.08774, 2023.

\bibitem{pang2023whitebox}
Yan Pang, Tianhao Wang, Xuhui Kang, Mengdi Huai, and Yang Zhang.
\newblock White-box membership inference attacks against diffusion models,
  2023.

\bibitem{pytorch}
Adam Paszke, Sam Gross, Soumith Chintala, and et~al.
\newblock Automatic differentiation in pytorch.
\newblock 2017.

\bibitem{sklearn}
Fabian Pedregosa, Ga{\"{e}}l Varoquaux, Alexandre Gramfort, and et~al.
\newblock Scikit-learn: Machine learning in python.
\newblock {\em J. Mach. Learn. Res.}, 12:2825--2830, 2011.

\bibitem{glove}
Jeffrey Pennington, Richard Socher, and Christopher~D. Manning.
\newblock Glove: Global vectors for word representation.
\newblock In {\em Proceedings of the {EMNLP}}, pages 1532--1543, 2014.

\bibitem{dps-fvae}
Bjarne Pfitzner and Bert Arnrich.
\newblock Dpd-fvae: Synthetic data generation using federated variational
  autoencoders with differentially-private decoder.
\newblock {\em CoRR}, abs/2211.11591, 2022.

\bibitem{DPPublicPretraining2}
Hieu Pham, Zihang Dai, Golnaz Ghiasi, and et~al.
\newblock Combined scaling for zero-shot transfer learning.
\newblock {\em CoRR}, abs/2111.10050, 2021.

\bibitem{dalle2}
Aditya Ramesh, Prafulla Dhariwal, Alex Nichol, and et~al.
\newblock Hierarchical text-conditional image generation with {CLIP} latents.
\newblock {\em CoRR}, abs/2204.06125, 2022.

\bibitem{StableDiffusion}
Robin Rombach, Andreas Blattmann, Dominik Lorenz, and et~al.
\newblock High-resolution image synthesis with latent diffusion models.
\newblock In {\em {IEEE/CVF} Conference on Computer Vision and Pattern
  Recognition, {CVPR}}, pages 10674--10685, 2022.

\bibitem{nolargemodel2}
Yinchen Shen, Zhiguo Wang, Ruoyu Sun, and v.
\newblock Towards understanding the impact of model size on differential
  private classification.
\newblock {\em CoRR}, abs/2111.13895, 2021.

\bibitem{ddim}
Jiaming Song, Chenlin Meng, and Stefano Ermon.
\newblock Denoising diffusion implicit models.
\newblock In {\em 9th International Conference on Learning Representations,
  {ICLR}}, 2021.

\bibitem{ddpm_song}
Yang Song and Stefano Ermon.
\newblock Generative modeling by estimating gradients of the data distribution.
\newblock In {\em Advances in Neural Information Processing Systems}, pages
  11895--11907, 2019.

\bibitem{Semantic4}
Robin Strudel, Ricardo~Garcia Pinel, Ivan Laptev, and et~al.
\newblock Segmenter: Transformer for semantic segmentation.
\newblock In {\em International Conference on Computer Vision}, pages
  7242--7252, 2021.

\bibitem{Semantic5}
Saeid~Asgari Taghanaki, Kumar Abhishek, Joseph~Paul Cohen, and et~al.
\newblock Deep semantic segmentation of natural and medical images: a review.
\newblock {\em Artif. Intell. Rev.}, 54(1):137--178, 2021.

\bibitem{dpsgd-gan8}
Reihaneh Torkzadehmahani, Peter Kairouz, and Benedict Paten.
\newblock {DP-CGAN:} differentially private synthetic data and label
  generation.
\newblock In {\em {IEEE} Conference on Computer Vision and Pattern Recognition
  Workshops, {CVPR} Workshops}, pages 98--104, 2019.

\bibitem{DPPublicPretraining1}
Florian Tram{\`{e}}r, Gautam Kamath, and Nicholas Carlini.
\newblock Considerations for differentially private learning with large-scale
  public pretraining.
\newblock {\em CoRR}, abs/2212.06470, 2022.

\bibitem{ImageCaption1}
Oriol Vinyals, Alexander Toshev, Samy Bengio, and et~al.
\newblock Show and tell: {A} neural image caption generator.
\newblock In {\em {IEEE} Conference on Computer Vision and Pattern
  Recognition}, pages 3156--3164, 2015.

\bibitem{fmnist}
Han Xiao, Kashif Rasul, and Roland Vollgraf.
\newblock Fashion-mnist: a novel image dataset for benchmarking machine
  learning algorithms.
\newblock {\em CoRR}, 2017.

\bibitem{ImageCaption2}
Kelvin Xu, Jimmy Ba, Ryan Kiros, and et~al.
\newblock Show, attend and tell: Neural image caption generation with visual
  attention.
\newblock In {\em Proceedings of the 32nd International Conference on Machine
  Learning}, volume~37 of {\em {JMLR} Workshop and Conference Proceedings},
  pages 2048--2057, 2015.

\bibitem{Semantic3}
Raymond~A. Yeh, Chen Chen, Teck{-}Yian Lim, and et~al.
\newblock Semantic image inpainting with deep generative models.
\newblock In {\em {IEEE} Conference on Computer Vision and Pattern
  Recognition}, pages 6882--6890, 2017.

\bibitem{dpmia}
Samuel Yeom, Irene Giacomelli, and et~al.
\newblock Privacy risk in machine learning: Analyzing the connection to
  overfitting.
\newblock In {\em 31st {IEEE} Computer Security Foundations Symposium, {CSF}},
  pages 268--282, 2018.

\bibitem{DBLP:conf/cvpr/YinMVAKM21}
Hongxu Yin, Arun Mallya, Arash Vahdat, and et~al.
\newblock See through gradients: Image batch recovery via gradinversion.
\newblock In {\em {IEEE} Conference on Computer Vision and Pattern Recognition,
  {CVPR}}, pages 16337--16346, 2021.

\bibitem{AwareDP2}
Da~Yu, Saurabh Naik, Arturs Backurs, and et~al.
\newblock Differentially private fine-tuning of language models.
\newblock In {\em The Tenth International Conference on Learning
  Representations, {ICLR}}, 2022.

\bibitem{dpsgd-gan9}
Xinyang Zhang, Shouling Ji, and Ting Wang.
\newblock Differentially private releasing via deep generative model.
\newblock {\em CoRR}, abs/1801.01594, 2018.

\bibitem{privsyn}
Zhikun Zhang, Tianhao Wang, Ninghui Li, and et~al.
\newblock Privsyn: Differentially private data synthesis.
\newblock In {\em 30th {USENIX} Security Symposium}, pages 929--946, 2021.

\bibitem{CVPretaining1}
Luowei Zhou, Hamid Palangi, Lei Zhang, and et~al.
\newblock Unified vision-language pre-training for image captioning and {VQA}.
\newblock In {\em The Thirty-Fourth {AAAI} Conference on Artificial
  Intelligence, {AAAI}}, pages 13041--13049, 2020.

\end{thebibliography}

\vspace{2pt}
\begin{center}
    \Large \bf Appendix
\end{center}

\setcounter{section}{0}
\setcounter{equation}{0}
\renewcommand\thesection{\Alph{section}}

\section{Selection of Privacy Budgets}
\label{ap:epsChoice}
DP provides a theoretical guarantee for data protection and has been adopted in many data synthesis tasks~\cite{dpgenreview}. The privacy parameter $\epsilon$ simultaneously affects the privacy strength and utility of the synthetic data. As $\epsilon$ decreases, more noise needs to be added to the clean gradient, which leads to a drop in the performance of our algorithm. As our experiments show, the classification accuracy (CA) of \toolname with $\epsilon=1$ is lower than that with $\epsilon=10$. Therefore, it could be useful to choose a proper $\epsilon$ value for some known attacks. In fact, this has been a rarely explored problem in the DP scenario, both theoretically and experimentally. As the definition of DP (Definition~\ref{def:dp}) shows, $\epsilon$ bounds the privacy loss in a probabilistic manner. Different attack approaches require different amounts of privacy loss for a successful attack, which is usually difficult to define.

For membership inference attacks (MIAs), Yeom et al.\cite{dpmia} bound the membership advantage for an $\epsilon$-DP algorithm to $e^{\epsilon}-1$, where a lower membership advantage indicates a lower attack success rate. In the case of data poisoning attacks, Ma et al.\cite{dppoison1} demonstrate that $0.1$-DP learners are resistant to data poisoning attacks when the adversary is only capable of poisoning a small number of items. Hong et al.\cite{dppoison2} discover that DP-SGD, even in configurations that do not provide meaningful privacy guarantees, enhances the model's robustness against data poisoning attacks.

\begin{figure}[t]
    \centering
    \includegraphics[width=0.9\linewidth]{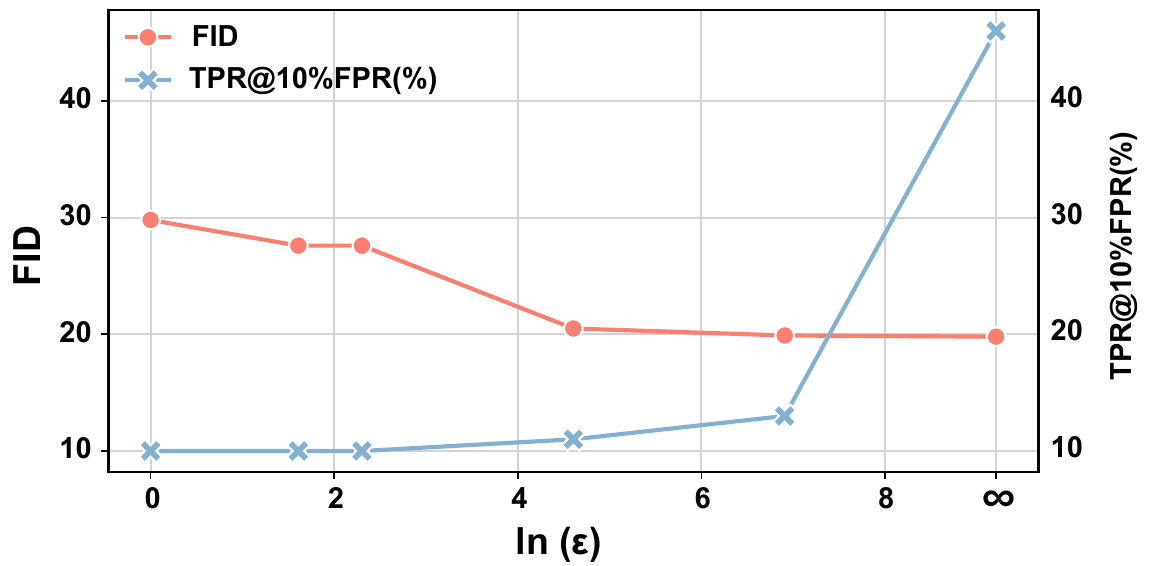}   
    \caption{The FID of synthetic images from \toolname and TPR@10\%FPR of MIA~\cite{diffusionAttacker3} under different $\epsilon$. "$\infty$" represents training \toolname without DP protection.}
    \label{ap:mia}
\end{figure}

To elaborate on what levels of DP are needed for \toolname to be resistant to known attacks and how this affects the utility of synthetic datasets, we choose a white-box MIA~\cite{diffusionAttacker3} for attacking diffusion models. We use the TPR@10\%FPR to evaluate the performance of the attacker. Specifically, TPR@10\%FPR refers to the True Positive Rate when the False Positive Rate is fixed at 10\%, and a higher metric means a higher attack success rate~\cite{diffusionAttacker3}. We study the vulnerability of diffusion models to MIA when trained with six different privacy budgets: $\epsilon \in \{1, 5, 10, 100, 1000, \infty\}$, where "$\infty$" represents training \toolname without DP protection. Figure~\ref{ap:mia} illustrates that as $\epsilon$ increases, \toolname presents a reduction in FID scores, signifying enhanced utility of the synthetic dataset. However, it is observed that an increased $\epsilon$ enables the attacker to attain a higher TPR@10\%FPR, suggesting that the data becomes more vulnerable to attacks. 
When $\epsilon$ is set to 100, the MIA achieves merely an 11\% TPR@10\%FPR, which approximates the effectiveness of random guessing. Consequently, within the context of the examined MIA, \toolname offers an effective defense mechanism. For practical applications, it is recommended to adopt smaller $\epsilon$ values, such as 10 or 1, as suggested by prior studies~\cite{privsyn,dpdm-sota,dpdm}, to defend against some unknown attacks.

\section{Semantics Query and Selection}
\label{ap:sqs}
In Section~\ref{subsec:qis}, we use a trained semantic query function to query $k_1$ semantics of each sensitive image. In Section~\ref{subsec:sdq}, we select the top-$k_2$ semantics based on their probabilities in the queried semantic distribution as the semantics description of the sensitive dataset. In our experiments, we set $k_1=k_2=k$. We take the example in Figure~\ref{fig:sdq} to tell the reason for this setting.

\noindent \textbf{(1) $k_1 < k_2$.} Since the number of semantics in the public dataset is 2, we have $k_1=1$ and $k_2=2$. The initial query results yield $\{zebra:3,bee:0\}$. However, we select both \textit{zebra} and \textit{bee} as the semantics description, meaning the entire public dataset is selected.

\noindent \textbf{(2) $k_1 > k_2$.} Since the number of semantics in the public dataset is 2, we have $k_1=2$ and $k_2=1$. The initial query results yield $\{zebra:3,bee:3\}$. Since the mean of added Gaussian noise is zero, it is random for us to select \textit{zebra} or \textit{bee} from the noisy semantic distribution as the semantics description of sensitive dataset.

Therefore, we set $k_1=k_2=k$ for our \toolname. Although the example from Figure~\ref{fig:sdq} is not very general. We find that this setting works well in our experiments and reduces the number of hyper-parameters of \toolname.

\section{Over-parameterized Models are Ineffective }
\label{ap:OMI}
We conduct experiments to verify that for DP image synthesis, where the gradient is noisy, fine-tuning an over-parameterized generative model on the sensitive dataset is not a proper way. Although previous studies have stated a similar view~\cite{dpdm,nolargemodel1,nolargemodel2}, none of them has conducted experiments to show that this phenomenon does exist in diffusion models. 

Specifically, we train diffusion models with different parameters 1.75M, 6.99M, 27.9M, 62.7M, 112M, 251M on {\tt MNIST}~\cite{mnist}. {\tt MNIST} contains 60,000 handwritten digits gray images, from 0 to 9, and has been used in more previous DP image synthesis studies~\cite{dpsgd-gan1,dp-merf}. The gradient used to update the models' parameters is added with Gaussian noise for satisfying ($10,1 \times {10^{ - 5}}$)-DP. We use the FID and CA of synthetic images from trained models to assess their synthesis quality. 

\begin{figure}[t]
    \centering
    \includegraphics[width=0.93\linewidth]{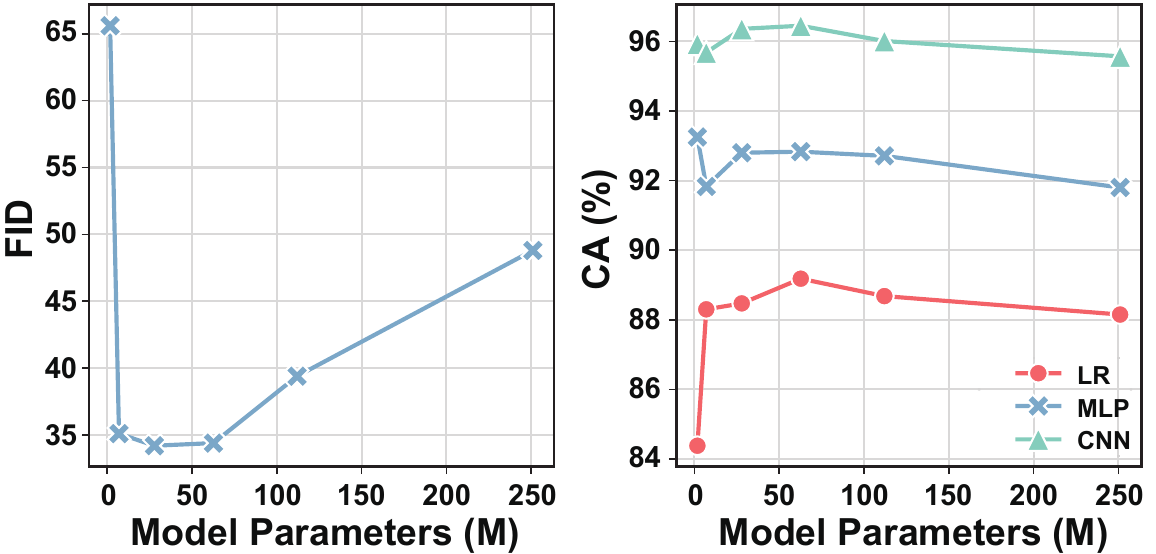}
    \caption{The FID and CA of the synthetic images from diffusion models of different sizes, which are fine-tuned on {\tt MNIST} with DP-SGD. }
    \label{ap:mnist_size}
\end{figure}

\begin{table*}[!t]
    \centering
    \caption{Hyper-parameters for training DPDM, PDP-Diffusion, DP-LDM and \toolname+D.}
    \label{ap:diffusion_details}
    \begin{tabular}{l|ccc|ccc}
    \toprule
    \multirow{2}{*}{Hyper-parameter}& \multicolumn{3}{c}{Pre-training} & \multicolumn{3}{c}{Fine-tuning}\\
    \Xcline{2-7}{0.5pt}
    & {\tt CIFAR-10}& {\tt CelebA32}& {\tt CelebA64} & {\tt CIFAR-10}& {\tt CelebA32}& {\tt CelebA64}\\
    \hline
    Learning rate & 1×$10^{-4}$& 1×$10^{-4}$& 1×$10^{-4}$ & 3×$10^{-4}$ & 3×$10^{-4}$ & 3×$10^{-4}$\\
    Training iterations & 128M& 42K & 42K & 152 & 618 & 1236 \\
    Batch size & 2048 & 2048 & 2048 & 16384 & 16384 & 8192\\
    Parameters & 1.75M & 1.80M & 2.00M & 1.75M & 1.80M & 2.00M\\
    \bottomrule
\end{tabular}
% \vspace{-8.0pt}
\end{table*}

Figure~\ref{ap:mnist_size} shows the results. When the parameters of diffusion models are very few (e.g. 1.75M), both FID and CA of the synthetic images are very poor. Because such few parameters are not enough for diffusion models to learn the distribution of the sensitive data, thus cannot generate images with high fidelity and utility. When the parameters are between 27.9M and 62.7M, the diffusion models seem to achieve optimal synthesis performance for the lowest FID and highest CA. However, when the parameters get more than 62.7M, both FID and CA get worse with the parameters increase. It is notable that the parameters of SOTA method are 80M. With such a number of parameters, the dimension of gradient gets large, which is called `curse of dimensionality'. As shown in Algorithm~\ref{alg:fine-tuning}, when we scale the gradient into a given maximal $l_2$-norm $C$, the scaled gradient added by Gaussian noise with the same variance typically becomes more noisy, leading more unstable training of diffusion models. Therefore, light models are more suitable for DP image synthesis.

\begin{figure}[!t]
\vspace{+3.0pt}
    \centering
    \setlength{\abovecaptionskip}{0pt}
    \includegraphics[width=1\linewidth]{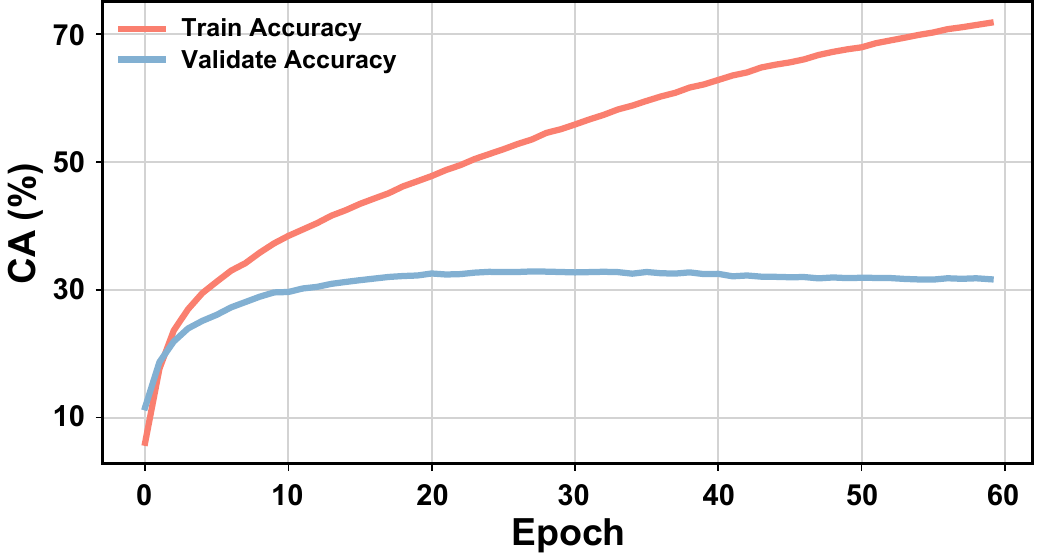}
    \caption{The semantic classification accuracy of the semantic query function on the training and validation set of {\tt ImageNet}~\cite{imagenet} during training. The best accuracy on the validation set is 32.9\%.}
    \label{fig:ap_sqf_acc}
\end{figure}

\section{Implementation Details}
\label{ap:ImpleDetails}

\subsection{Investigated Dataset}
\label{ap:dataset}

\begin{figure}[t]
    \centering
    \setlength{\abovecaptionskip}{0pt}
    \includegraphics[width=0.96\linewidth]{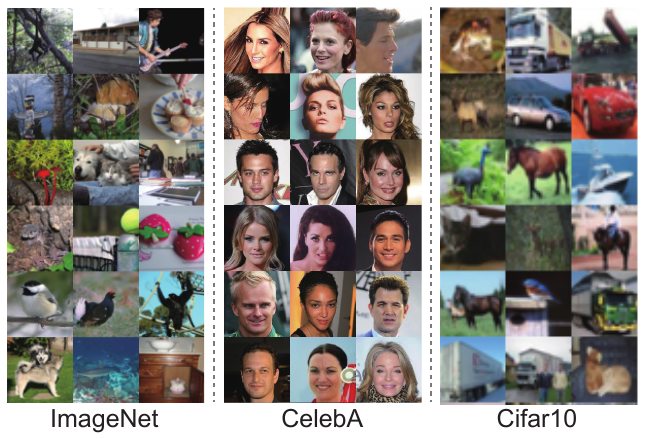}
    \caption{Images randomly sampled from the {\tt ImageNet}, {\tt CelebA}, and {\tt CIFAR-10} dataset.}
    \label{fig:dataset_samples}
\end{figure}

We utilize {\tt ImageNet}~\cite{imagenet}, a dataset commonly adopted for pre-training in computer vision venues, as our public dataset. Besides, we use {\tt CelebA}~\cite{celeba} and {\tt CIFAR-10}~\cite{cifar10} datasets, which are prevalent in the DP image synthesis domain~\cite{dpdm-sota,dp-mepf,api}, as the sensitive datasets. Compared to {\tt MNIST}~\cite{mnist} and {\tt FashionMNIST}~\cite{fmnist}, which are used more commonly in previous research, these two datasets are more difficult to synthesize. {\tt CelebA} is a large-scale face attributes dataset with over 200K celebrity images. The images in this dataset cover large pose variations and background clutter. The {\tt CIFAR-10} dataset comprises 60,000 natural images with a resolution 32×32, distributed across 10 unique classes, each containing 6,000 images. We present the examples of dataset in Figure~\ref{fig:dataset_samples}.

\subsection{Hyper-parameters}
We present the hyper-parameters of training Diffusion-based methods, DPDM, PDP-Diffusion and \toolnameD in Table~\ref{ap:diffusion_details}. Since we find that semantics queried from {\tt CIFAR-10} are more colorful than that from {\tt CelebA}, the pre-training iterations on {\tt CIFAR-10} are set more than that on {\tt CelebA}. For sampling, all diffusion model uses 50 steps and DDIM sampler. We present the hyper-parameters of training GAN-based methods, DPGAN, DPGAN-p and \toolnameG in Table~\ref{ap:gan_details}. The ‘update ratio’ represents the time we update the parameters of $Dis$ when we update the parameters of $Gen$ 1 time. 5 is a typical choice from previous studies~\cite{gan,biggan,wgan}. All Diffusion-based and GAN-based methods are employed with the same hyper-parameters on the same dataset for fair comparison. 

\begin{table*}[t]
    \centering
    \caption{Hyper-parameters for training DPGAN, DPGAN-p and \toolname+G.}
    \label{ap:gan_details}
    \begin{tabular}{l|ccc|ccc}
    \toprule
    \multirow{2}{*}{Hyper-parameter}& \multicolumn{3}{c}{Pre-training} & \multicolumn{3}{c}{Fine-tuning}\\
    \Xcline{2-7}{0.5pt}
    & {\tt CIFAR-10}& {\tt CelebA32}& {\tt CelebA64} & {\tt CIFAR-10}& {\tt CelebA32}& {\tt CelebA64}\\
    \hline
    $Gen$ Learning rate & 2×$10^{-4}$ & 2×$10^{-4}$ & 2×$10^{-4}$& 2×$10^{-4}$ & 2×$10^{-4}$ & 2×$10^{-4}$\\
    $Dis$ Learning rate & 2×$10^{-4}$ & 2×$10^{-4}$ & 2×$10^{-4}$& 2×$10^{-4}$ & 2×$10^{-4}$ & 5×$10^{-4}$\\
    Update ratio& 5& 5& 5& 5& 5& 2\\
    Training iterations & 20K & 4K & 4K & 30 & 124 & 124\\
    Batch size & 64 & 64 & 64 & 16384 & 16384 & 16384\\
    Parameters of $Gen$& 9.42M & 9.42M & 11.3M& 9.42M & 9.42M & 11.3M\\
    Parameters of $Dis$& 2.37M & 2.37M & 19.6M& 2.37M & 2.37M & 19.6M\\
    \bottomrule
\end{tabular}
% \vspace{-8.0pt}
\end{table*}

\begin{table}[t]
% \vspace{-7.0pt}
    \centering
    \caption{Hyper-parameters for semantic query function. The scheduler indicates how we adjust the learning rate during the classifier's training.}
    \label{ap:sqf_hyperparam}
    \begin{tabular}{l|c}
    \toprule
    Hyper-parameter& Setting\\
    \hline
    Learning rate& 0.01\\
    Training iterations & 9.4K\\
    Batch size & 8192\\
    Scheduler & CosineAnnealingLR\\
    \bottomrule
\end{tabular}
% \vspace{-4.0pt}
\end{table}

\subsection{Details of \toolname}
Different from existing methods, \toolname train a semantic query function and use it to obtain the semantic distribution of the sensitive dataset without leaking privacy. Therefore, we divide this section into details of semantic query function and semantic distribution query.

\noindent \textbf{Semantic Query Function.} We use ResNet50~\cite{resnet} to implement our semantic query function and the code is provided in the package PyTorch~\cite{pytorch}. We train the semantic query function on the training set of {\tt ImageNet} ~\cite{imagenet} with the objective function (Eq.~\eqref{eq:L_Q}) and use its classification accuracy on the validation set to determine whether stopping the training. Since the minimal resolution of sensitive images is 32×32, we resize all the images into 32×32 for training. The classification accuracy curves of trained semantic query function on the training set and validation set are showed in Figure~\ref{fig:ap_sqf_acc} and more training details are showed in Table~\ref{ap:sqf_hyperparam}.

\noindent \textbf{Semantic Distribution Query.}  There are two hyper-parameters in our Semantic Distribution Query, the number of queried semantics for each sensitive image $k_1$, the number of selected semantics from the queried semantic distribution $k_2$ and the variance $\sigma_2^2$ of Gaussian noise added to the query result $SD$. In Appendix~\ref{ap:sqs}, we set $k_1=k_2=k$.

For $k$ (e.g., selection ratio), we have explored how it affects our \toolname in RQ3 of Section~\ref{sec:rq}. We show the detailed setting of $k$ for different selection ratios on {\tt CIFAR-10} and {\tt CelebA} in Table~\ref{ap:k}. It is noticed that for the same selection ratio, the $k$ for {\tt CIFAR-10} is always more than that for {\tt CelebA}. That is because when there are multiple categories in the sensitive dataset, we usually hope to achieve conditional synthesis (e.g., generating a subdataset for each category). Thus, the synthetic images can be used to train a classifier. In this situation, we consider it more effective to represent the sensitive semantic distribution as a condition distribution. Specifically, we can select $k$ semantics for each category in the sensitive dataset, and each category has its own pre-training dataset selected by its $k$ semantics.

For $\sigma_2^2$, in our experiments, we find that the semantic distribution query works well even under a tiny privacy budget (e.g. 0.01 or 0.001). Therefore, the privacy budget allocated to fine-tuning is almost unchanged and the stability of fine-tuning is not harmed a lot. We set $\sigma_2$ as 484 and 5300 for {\tt CIFAR-10} and {\tt CelebA} respectively.

\subsection{Details of Classifiers}
We choose three classifiers LR, MLP and CNN to assess the utility of synthetic images on {\tt CIFAR-10}.  The LR and MLP are from scikit-learn~\cite{sklearn} with the default hyper-parameters. For CNN, we use ResNet9 taken from FFCV~\cite{ffcv}.

\section{Missing Proofs}
\label{ap:MissingProofs}
\textbf{Proof of Theorem~\ref{the:sd}: The semantic distribution $SD$ has global sensitivity $\Delta_{SD}=\sqrt{k_1}$, where $k_1$ is the number of semantics we query from each sensitive image. For any $\alpha>1$, incorporating noise $\mathcal{N}\left( {0,k_1{\sigma_2 ^2} {\rm I}} \right)$ into the semantic distribution $SD$ makes the query results satisfies ($\alpha$, $\alpha$/($2\sigma_2^2$))-RDP.}

\textit{Proof.} We first prove the semantic distribution $SD$ has global sensitivity $\Delta_{SD}=\sqrt{k_1}$.  A semantic distribution $SD$ specified by a set of semantics $S$ is a frequency distribution table, showing the number of images with each semantic existing in the public dataset. For two semantic distributions $SD_0$ and $SD_1$, where $SD_1$ is obtained by adding or removing one image to $SD_0$. In general, for any $k_1$, it is obviously that ${\Delta _{SD}} = {\left\| {S{D_0} - S{D_1}} \right\|_2} = \sqrt k_1 $.

Let $SD + N\left( {0,k_1\sigma _2^2{\rm{I}}} \right) = \sqrt{k_1} \left( {SD/ \sqrt{k_1} + N\left( {0,\sigma _2^2{\rm{I}}} \right)} \right)$. Similar to above proof, $SD/ \sqrt{k_1}$ has the global sensitivity ${\Delta _{SD/\sqrt{k_1}}} = 1$. According to the Corollary 3 from~\cite{rdp}, we have ${SD/ \sqrt{k_1} + N\left( {0,\sigma _2^2{\rm{I}}} \right)}$ satisfies ($\alpha$, $\alpha$/($2\sigma_2^2$))-RDP. Therefore, $SD + N\left( {0,k_1\sigma _2^2{\rm{I}}} \right)$ also  satisfies ($\alpha$, $\alpha$/($2\sigma_2^2$))-RDP.

\begin{table}[t]
% \vspace{-7.0pt}
    \centering
    \caption{The choice of $k$ for different selection ratios on {\tt CIFAR-10} and {\tt CelebA}. {\tt CIFAR-10} has 10 categories and $k=1$ means that we need select 10 subsets from {\tt ImageNet}.}
    \label{ap:k}
    \begin{tabular}{c|cc}
    \toprule
    Selection ratio& {\tt CIFAR-10} &{\tt CelebA}\\
    \hline
    1\%& 1 & 10\\
    5\% & 5 & 50\\
    10\% & 10 & 100\\
    20\% & 20 & 200\\
    50\% & 50 & 500\\
    \bottomrule
\end{tabular}
% \vspace{-8.0pt}
\end{table}

\begin{figure*}[!t]
    \centering
    % \vspace{+5.0pt}
    % \setlength{\abovecaptionskip}{0pt}
    \includegraphics[width=0.95\linewidth]{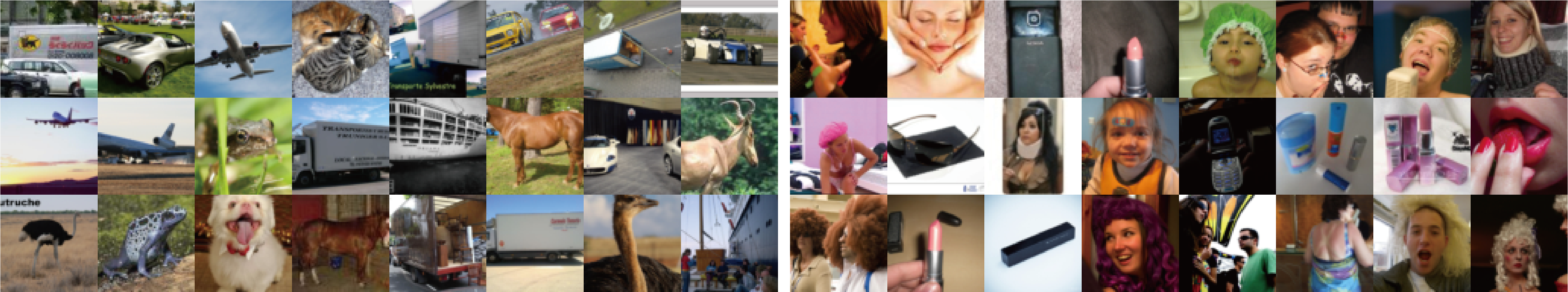}
    \caption{(Left) Samples from selected dataset for {\tt CIFAR-10} with top-10 semantics: “airline”, “sport car”, “ostrich”, “tabby”, “hartebeest”, “pekinese”, “tailed frog”, “sorrel”, “ocean liner” and “moving van”. (Right) Samples from selected dataset for {\tt CelebA} with top-10 semantics: “wig”, “neck brace”, “band aid”, “lipstick”, “sunglasses”, “sunglass”, “mobile phone”, “hair slide”, “sunscreen” and “shower cap”.}
    \label{fig:visualSimi}
\end{figure*}

\section{Privacy Analysis}
\label{ap:pa}
In \toolname, two processes consume the privacy budget: (1) querying the semantic distribution of the sensitive dataset and (2) querying the gradient during each training iteration while fine-tuning on the sensitive dataset. Both these processes can be viewed as compositions of multiple Sub-sampled Gaussian Mechanism (SGM) instances~\cite{sgm}. As explained in Theorem~\ref{the:rdp4sgm}, we utilize RDP to monitor the privacy cost throughout these processes. 
RDP is advantageous for deep learning methods that involve many iterations~\cite{dpsgd,sgm}. When training a deep neural network with differential privacy, each iteration might introduce some privacy loss. Through leveraging RDP, as explained in Theorem~\ref{compositionRDP}, we can accurately track and control these losses, ensuring overall privacy protection remains within acceptable boundaries.

\begin{theorem}
\label{compositionRDP}
    (Composition by RDP~\cite{sgm}) For a sequential of $T$ SGM ${M_1}, \ldots ,{M_T}$  satisfying  ($\alpha, \gamma_i$)-RDP (Definition~\ref{def:rdp}) for $i = 1, \ldots ,T$, any $\alpha>1$ and $\delta>0$, the sequential composition satisfies ($\varepsilon, \delta$)-DP and $\varepsilon  = \mathop {\min }\limits_\alpha  \left( {\sum\limits_{i = 1}^T {{\gamma _i(\alpha,\sigma_i,q_i)}}  - \log \left( \delta  \right)/\left( {\alpha  - 1} \right)} \right)$.
\end{theorem}
\noindent The ${\gamma _i(\alpha,\sigma_i,q_i)}$ is related to the $i$-th SGMs with the variance of Gaussian noise $\sigma_i^2$, sample ratio $q_i$. Theorem~\ref{compositionRDP} presents that the final result of \toolname for ($\varepsilon, \delta$)-DP can be obtained by optimizing over $\alpha$. Since the sample ratio (or batch size) during the fine-tuning is usually fixed, we have the following Theorem of composition by RDP for \toolname,
\begin{theorem}
\label{the:compositionRDP}
    (\toolname Privacy Composition) For given $\delta>0$, \toolname composed of $T$ iterations of fine-tuning with the variance of Gaussian noise $\sigma_1^2$, combined with an one-shot semantic distribution query with the variance of Gaussian noise $\sigma_2^2$ satisfies ($\varepsilon, \delta$)-DP. The value of $\varepsilon$ is determined by $\varepsilon  = \mathop {\min }\limits_\alpha  \left( {T{\gamma _F}\left( {\alpha ,{\sigma _1},q} \right) + \alpha /\left( {2\sigma _2^2} \right) - \log \left( \delta  \right)/\left( {\alpha  - 1} \right)} \right)$.
\end{theorem}

\noindent ${{\gamma _F}\left( \alpha,\sigma_1,q  \right)}$ and ${{\gamma _S}\left( \alpha,\sigma_2,1  \right)}$ are related to querying the gradient and semantic distribution respectively, which can be calculated following Theorem~\ref{the:rdp4sgm}.  Typically, we set the training iterations $T_m$ and sample ratio (or batch size) $q$ before we train our models. Given sensitive images sampling ratio $q$, training iterations $T_m$, Gaussian variance $\sigma_2^2$ for semantic distribution query and privacy budget ($\varepsilon,\delta$), the standard deviation $\sigma_1$ can be obtained through the following theorem.
\begin{theorem}
\label{the:obtainSigma}
    Given sensitive images sampling ratio $q$, training iterations $T_m$ and Gaussian variance $\sigma_2^2$ for semantic distribution query, \toolname satisfies ($\varepsilon, \delta$)-DP with Gaussian variance $\sigma_1^2$ for gradient sanitizing, where $\sigma_1 = \mathop {\arg \min }\limits_{{\sigma _1}} {\left[ {\varepsilon  - \mathop {\min }\limits_\alpha  \left( {T_m {\gamma _F}\left( {\alpha ,{\sigma _1},q} \right) + \frac{\alpha }{{2\sigma _2^2}} - \frac{{\log \left( \delta  \right)}}{{\alpha  - 1}}} \right)} \right]^2}$.
\end{theorem}

With Theorem~\ref{the:obtainSigma}, we can find a proper $\sigma_2$ to make our \toolname satisfy given ($\varepsilon, \delta$)-DP through Theorem~\ref{the:obtainSigma}. We achieve this via the RDP accountant tool provided by opacus. In particular, we randomly initialize a list of $\sigma_2$ to calculate their $\varepsilon$ through Theorem~\ref{the:compositionRDP}. We can find a $\sigma_2$, whose $\varepsilon$ is closest to our target privacy budget. We then generate a new list whose mean is the last optimal $\sigma_2$. Repeat the above process until we find an $\varepsilon$ satisfying our requirement. 

\begin{figure*}[!t]
    \centering
    % \vspace{+5.0pt}
    % \setlength{\abovecaptionskip}{0pt}
    \includegraphics[width=1\linewidth]{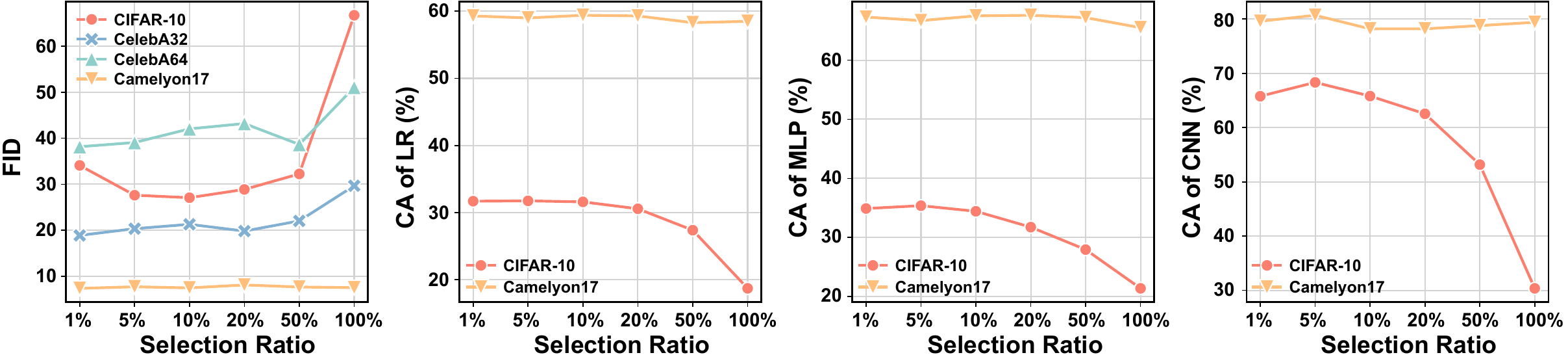}
    \caption{The effect of the selection ratio on the FID of the synthetic image dataset and the
Classification Accuracy for downstream classification tasks on different datasets.}
    \label{fig:ap_camelyon17}
    \vspace{+8.0pt}
\end{figure*}

\section{More Results}

\subsection{Larger Domain Shifts}
\label{ap:camelyon17}

\begin{figure*}[t]
    \scriptsize
    \centering
    \begin{minipage}[t]{\linewidth}
    
    \begin{minipage}[t]{0.5\linewidth}
        \centering
        \centerline{\includegraphics[width=1\linewidth]{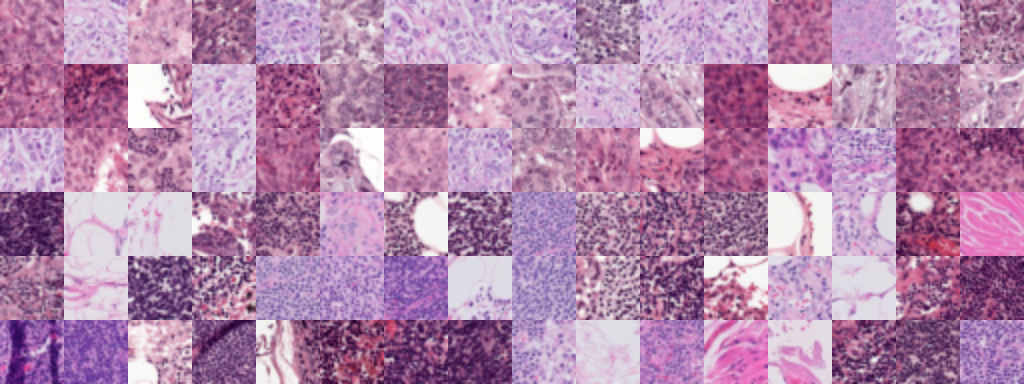}}
        \centerline{\large Real}
    \end{minipage}
    \begin{minipage}[t]{0.5\linewidth}
        \centering
        \centerline{\includegraphics[width=1\linewidth]{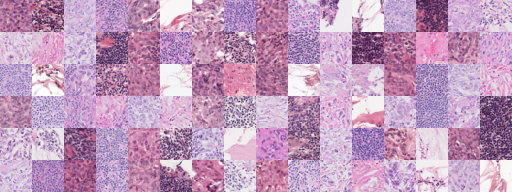}}
        \centerline{\large Synthetic}
    \end{minipage}
        
    \end{minipage}
    \caption{The left are real samples from {\tt Camelyon17}. The right are synthetic images from \toolname pre-trained with only 1\% images of {\tt ImageNet}. The privacy budget is $(10,10^{-6})$.}
    \label{fig:camelyonSyn}
    \vspace{+5.0pt}
\end{figure*}

In RQ3, we find that \toolname performs better when the selection ratio is smaller, an effect that is less pronounced on {\tt CelebA} compared to {\tt CIFAR-10} due to the larger domain shift. To further explore this phenomenon, we conduct an ablation study on a more "sensitive" dataset, {\tt Camelyon17}~\cite{camelyon1,camelyon2}, which comprises 455,954 histopathological 96 × 96 image patches of human tissue. {\tt Camelyon17} is less similar to {\tt ImageNet}, which is composed of natural images, than {\tt CIFAR-10} and {\tt CelebA}. We choose the same selection ratios as in RQ3 and calculate the FID and CA of synthetic images from \toolname under different selection ratios. Figure~\ref{fig:ap_camelyon17} shows that, on {\tt Camelyon17}, both FID and CAs of three different classifiers are almost unaffected by the selection ratio, in contrast to the results on {\tt CIFAR-10} and {\tt CelebA}. When the distribution discrepancy between the sensitive and public datasets is large, it is difficult to select useful data for pre-training, and the benefit of the selective data to pre-training is minimal. However, even on such a "sensitive" dataset, our selective data for pre-training does not lead to worse performance compared to using the entire {\tt ImageNet} for pre-training. When the available public dataset is substantially different from our sensitive dataset, we can use the semantic query function to decrease the amount of data for pre-training, which can greatly save computational resources.

We also provide some real samples and synthetic samples from \toolname pre-trained with only 1\% of ImageNet on Camelyon17 in Figure~\ref{fig:ap_camelyon17}.

\subsection{Visual Similarity}
\label{ap:visualSimi}

We show samples of the dataset selected by our \toolname with queried semantic distribution in Figure~\ref{fig:visualSimi}. The selection ratio for {\tt CIFAR-10} and {\tt CelebA} are 1\% and 5\% respectively. 

For {\tt CIFAR-10}~\cite{cifar10}, our \toolname selects useful semantics, such as “ostrich” and “tabby” whose images are usually similar to bird's and cat's respectively. For {\tt CelebA}~\cite{celeba}, which differs a lot from the public dataset, our \toolname also selects useful semantics, like “sunglasses” and “hair slide” whose images may contain human face in some area. Using these images to pre-train image generative models can guide models better to concern about the image information more related to the sensitive data without leaking privacy. 
% \gc{DPSDA is capable of generating highly detailed facial images, yet the consistency of its sampling quality remains variable.}

\subsection{More Synthetic Images}
\label{ap:moreSamples}

\begin{figure*}[!t]
    \centering
    \setlength{\abovecaptionskip}{-5pt}
    \includegraphics[width=1\linewidth]{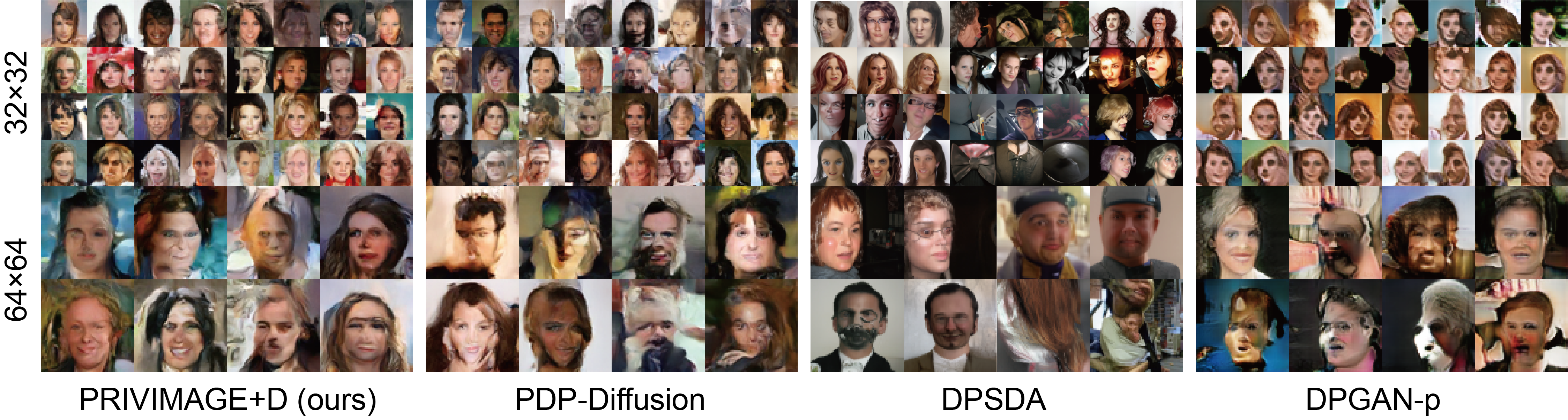}
    \caption{Examples of Synthetic {\tt CelebA32} and {\tt CelebA64}~\cite{celeba} images generated by \toolnameD (ours), PDP-Diffusion~\cite{dpdm-sota}, DPSDA~\cite{api} and DPGAN-p, respectively.}
    \label{fig:celeba_comp_show}
    \vspace{-20.0pt}
\end{figure*}

Examples of synthetic images from various methods on {\tt CIFAR-10}, {\tt CelebA32}, and {\tt CelebA64} are presents in Figure~\ref{fig:cifar10_comp_show} and Figure~\ref{fig:celeba_comp_show}. 
% Due to space limitation, we put the Figure~\ref{fig:celeba_comp_show} in Appendix~\ref{ap:moreSamples}.
For synthetic {\tt CIFAR-10} images, \toolname generates images with clear distinctions between different categories. Synthetic images produced by PDP-Diffusion~\cite{dpdm-sota} are indistinguishable across different categories, resulting in classifiers trained on these images having poor classification accuracy. While DPGAN-p can produce synthetic {\tt CIFAR-10} images that distinctly represent different categories, their generation quality is inconsistent, as shown in the first row, first column, and first row, fourth column of Figure~\ref{fig:cifar10_comp_show}. These low-quality images adversely affect the performance of classifiers trained on the synthetic data. The visual quality of the synthetic images generated by DPSDA is comparable to that of \toolname. However, DPSDA shares the same issue as DPGAN-p, as shown in the second row, second column. These anomalous samples could potentially reduce the performance of trained classifiers.

% We provide more examples of synthetic images from our \toolname with privacy budget ($\varepsilon=10,\delta=1 \times 10^{-6}$). 

For {\tt CelebA}, Figure~\ref{fig:celeba_comp_show} shows the synthetic images of {\tt CelebA32} and {\tt CelebA64} generated by \toolnameD, PDP-Diffusion . DPSDA and DPGAN-p. Compared to PDP-Diffusion and DPGAN-p, \toolname generates synthetic face images with clearer facial contours and facial expression, indicating that our synthetic images are of more fidelity. For example, from our synthetic images with $64\times64$ resolution, we can see a clearer mouth and smiling expression.

\begin{figure*}[!t]
    \centering
    \setlength{\abovecaptionskip}{-5pt}
    \includegraphics[width=1\linewidth]{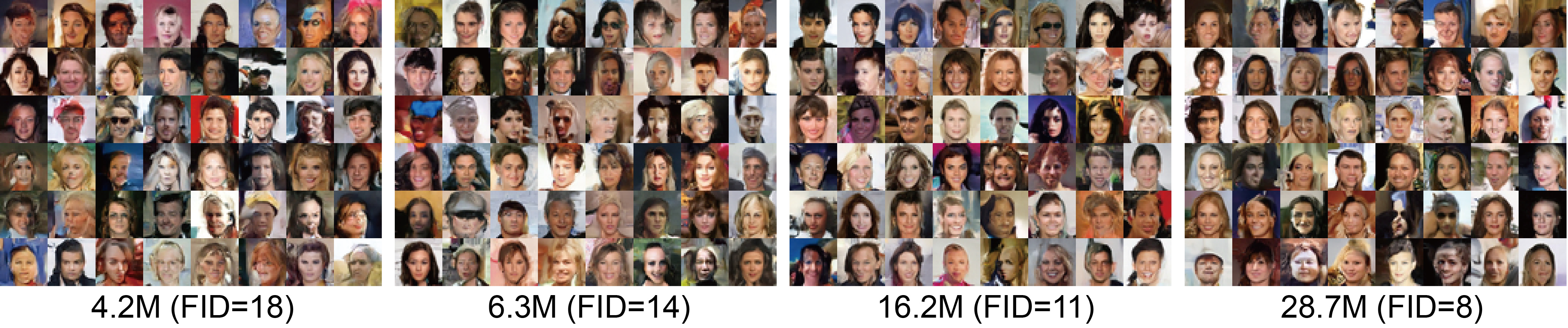}
    \caption{Examples of Synthetic {\tt CelebA32} images generated by \toolnameD with four different parameters.}
    \label{fig:more_celeba32}
    \vspace{-20.0pt}
\end{figure*}

\begin{figure*}[!t]
    \centering
    \setlength{\abovecaptionskip}{-5pt}
    \includegraphics[width=1\linewidth]{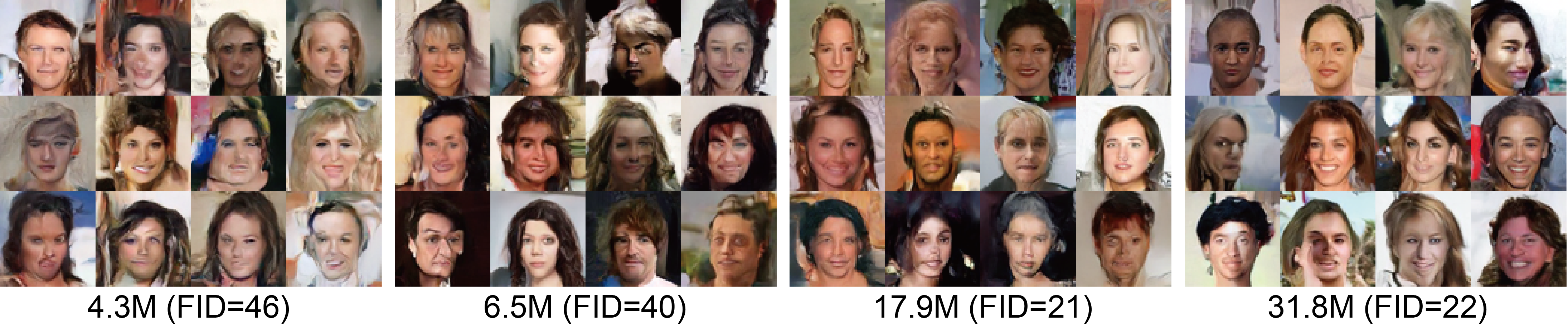}
    \caption{Examples of Synthetic {\tt CelebA64}~\cite{celeba} images generated by \toolnameD with four different parameters.}
    \label{fig:more_celeba64}
    \vspace{-20.0pt}
\end{figure*}

\begin{figure*}[!t]
    \centering
    \setlength{\abovecaptionskip}{-5pt}
    \includegraphics[width=1\linewidth]{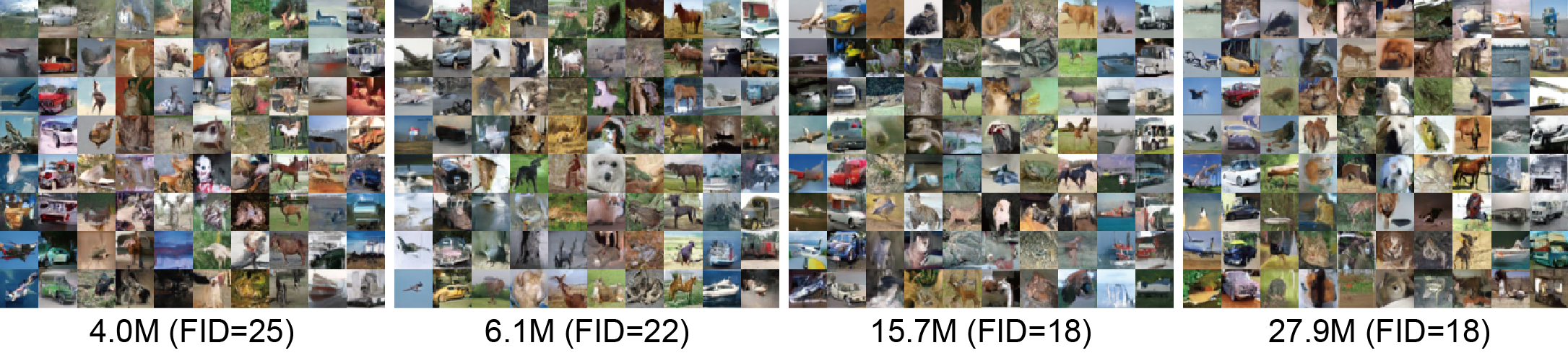}
    \caption{Examples of Synthetic {\tt CIFAR-10} images generated by \toolnameD with four different parameters. Each line corresponds to a category from the {\tt CIFAR-10} dataset.}
    \label{fig:more_cifar10}
    \vspace{-20.0pt}
\end{figure*}

Figures~\ref{fig:more_celeba32},~\ref{fig:more_celeba64} and~\ref{fig:more_cifar10} show the synthetic images of {\tt CelebA} and {\tt CIFAR-10} from our \toolnameD with different number of parameters as described in RQ4 of Section~\ref{sec:rq}. In Figure~\ref{fig:more_cifar10}, each line indicates a category from the {\tt CIFAR-10} dataset. With more parameters, \toolname can generate synthetic images of more fidelity.


\setcounter{section}{0}
\renewcommand\thesection{\Alph{section}}

\end{document}